\pdfoutput=1

\documentclass[11pt]{article}

\usepackage[preprint]{acl}

\usepackage{times}
\usepackage{latexsym}

\usepackage[T1]{fontenc}

\usepackage[utf8]{inputenc}

\usepackage{microtype}

\usepackage{inconsolata}

\usepackage{graphicx}

\usepackage{bm}
\usepackage{amsmath}
\usepackage{amssymb}
\usepackage{xspace}
\usepackage{graphicx}
\usepackage{multirow}
\usepackage{booktabs}
\usepackage{paralist}

\newcommand{\ie}{\emph{i.e.,}\xspace}
\newcommand{\eg}{\emph{e.g.,}\xspace}

\newcommand{\resp}{\emph{resp.}\xspace}

\newcommand{\modelname}{mGTE}

\usepackage{CJKutf8}

\title{mGTE: Generalized Long-Context Text Representation and Reranking Models for Multilingual Text Retrieval}

\author{
Xin Zhang\textsuperscript{\rm 1,2},
Yanzhao Zhang\textsuperscript{\rm 1},
Dingkun Long\textsuperscript{\rm 1},
Wen Xie\textsuperscript{\rm 1},
Ziqi Dai\textsuperscript{\rm 1},
\\ \textbf{
Jialong Tang\textsuperscript{\rm 1},
Huan Lin\textsuperscript{\rm 1}
Baosong Yang\textsuperscript{\rm 1},
Pengjun Xie\textsuperscript{\rm 1},
Fei Huang\textsuperscript{\rm 1},
}\\ \textbf{
Meishan Zhang\thanks{Corresponding Author},
Wenjie Li\textsuperscript{\rm 2},
Min Zhang
}\\
\textsuperscript{\rm 1}Tongyi Lab, Alibaba Group,
\textsuperscript{\rm 2}The Hong Kong Polytechnic University
\\
\texttt{\{linzhang.zx,zhangyanzhao.zyz,dingkun.ldk\}@alibaba-inc.com}
}

\begin{document}
\begin{CJK}{UTF8}{gbsn}

\maketitle
\begin{abstract}
We present systematic efforts in building long-context multilingual text representation model (TRM) and reranker from scratch for text retrieval.
We first introduce a text encoder (base size) enhanced with RoPE and unpadding, pre-trained in a native 8192-token context (longer than 512 of previous multilingual encoders).
Then we construct a hybrid TRM and a cross-encoder reranker by contrastive learning.
Evaluations show that our text encoder outperforms the same-sized previous state-of-the-art XLM-R.
Meanwhile, our TRM and reranker match the performance of large-sized state-of-the-art BGE-M3 models and achieve better results on long-context retrieval benchmarks.
Further analysis demonstrate that our proposed models exhibit higher efficiency during both training and inference.
We believe their efficiency and effectiveness could benefit various researches and industrial applications.\footnote{
Models are released at \url{https://hf.co/Alibaba-NLP/gte-multilingual-base}.
}
\end{abstract}

\section{Introduction}

Text retrieval aims to find relevant passages or documents from a large corpus given a query \cite{manning2008introduction}.
It is often implemented as a multi-stage process, consisting of two main components: a \emph{retriever} and a \emph{reranker} \cite{gao2021rethink,zhang2022hlatr,zhao2024dense}.
The retriever identifies a set of candidate documents that are potentially relevant to the query based on the similarity between their sparse (lexical term weights) or/and dense representations from a text representation model (TRM).
While the reranker reorders these retrieved candidates to refine the results based on the relevance score generated by a more precise yet computationally demanding model that processes both the query and a candidate document together.

Recent advances in large language models (LLMs) and retrieval augmented generation (RAG) \cite{gao2023retrieval} systems have led to an unprecedented surge in demand for versatile, plug-and-play TRMs and rerankers.
These new applications heavily involve processing long and multilingual texts, which could not be addressed by conventional encoder-based models and urgently require upgraded ones.
To this end, some resort to enhancing existing multilingual encoders, \eg XLM-R \cite{conneau-etal-2020-unsupervised}, with extended context window up to 8192 \cite{chen-etal-2024-m3}.
Others turn to use multilingual LLMs which already have the required capabilities \cite{zhang2023language}, but their models might be computationally expensive for self-hosted search services.

\begin{figure}[t]
\includegraphics[width=\columnwidth]{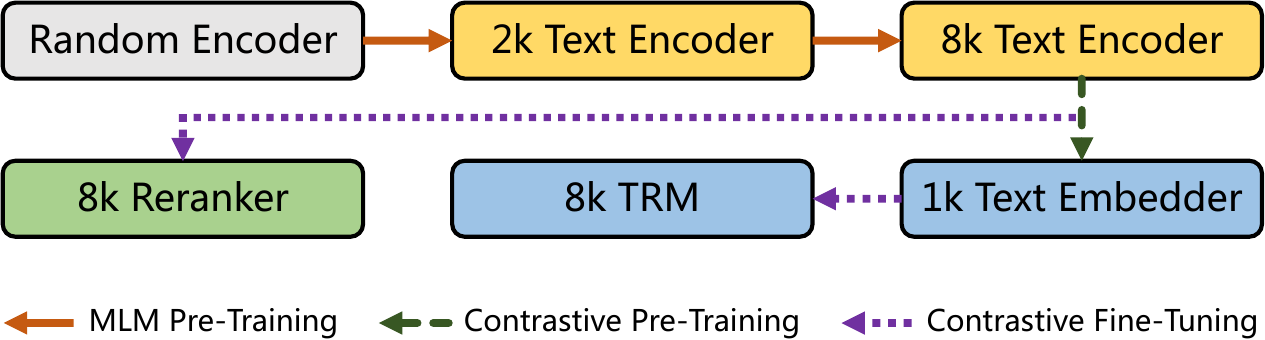}
\caption{
Training pipeline. We first build an 8k long-context multilingual encoder.
Then based on it, we train text representation and reranking models for retrieval.
}
\label{fig:pipeline}
\end{figure}

In the English community, it has been proven that training long-context encoders from scratch is promising for text retrieval  \cite{gunther2023jina2,nussbaum2024nomic}.
In this work, we continue this journey, presenting systematic efforts in building the long-context multilingual text encoder, TRM, and reranker.
We suggest a holistic pipeline (Figure \ref{fig:pipeline}) as well as several techniques in modeling and training for multilingual long-context retrieval.

Concretely, we first introduce a text encoder enhanced with Rotary Position Embedding (RoPE, \citealp{su2024roformer}) and unpadding \cite{portes2023mosaicbert}, pre-trained by masked language modeling (MLM) \cite{devlin-etal-2019-bert} via a two-stage curriculum for the native 8,192 tokens context.
Based on our encoder, we propose a hybrid TRM capable of generating both elastic dense \cite{kusupati2022matryoshka} and sparse vectors for efficient first-stage retrieval, as well as a cross-encoder reranker.
We construct them via the contrastive learning objective \cite{wang2022text,li2023towards} with large-scale meticulously curated datasets, providing robust off-the-shelf retrieval models.

We conduct extensive experiments to verify our method.
For the text encoder, we evaluate on two natural language understanding (NLU) benchmarks, \ie XTREME-R \cite{ruder-etal-2021-xtreme} and GLUE \cite{wang2018glue}, and show that our encoder outperforms the same-sized previous state-of-the-art XLM-R.
For the TRM and reranker, we evaluate on multiple retrieval benchmarks with multilingual and long-context settings, \eg MIRACL \cite{10.1162/tacl_a_00595} and MLDR \cite{chen-etal-2024-m3}, where our models match the performance of state-of-the-art BGE-M3 \cite{chen-etal-2024-m3} and achieve better long-context performance by a smaller size.
We open-source our models and code to facilitate further research and applications.

\begin{figure}\centering
\includegraphics[width=\columnwidth]{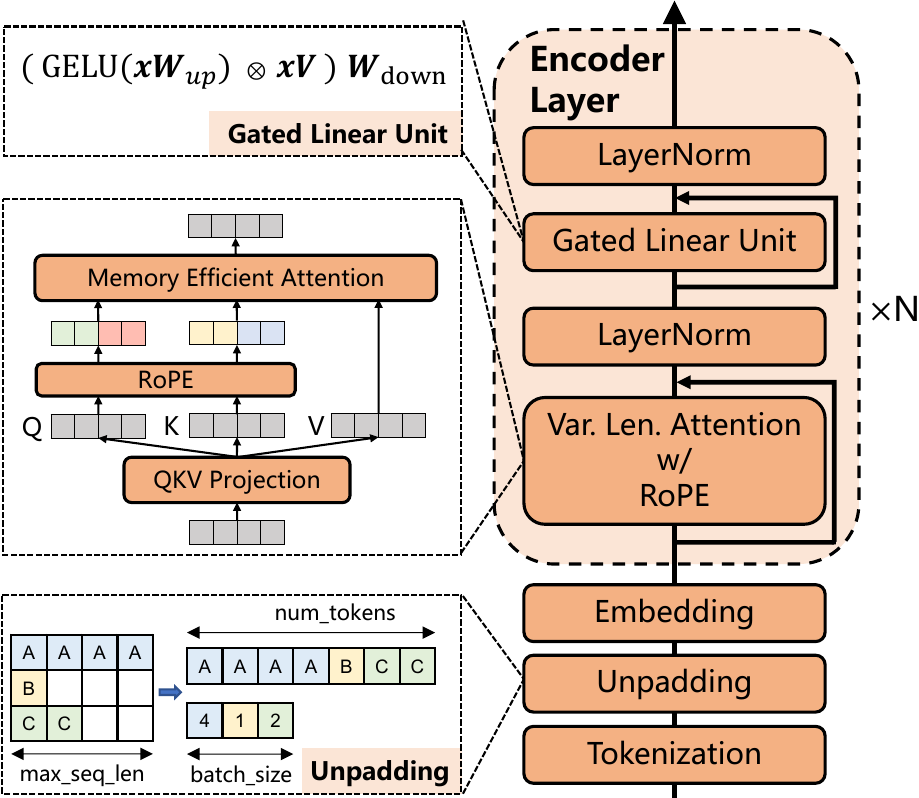}
\caption{
Our text encoder architecture.
}
\label{fig:encoder}
\end{figure}

\section{Method}

\subsection{Text Encoder}\label{sec:method:encoder}

To construct powerful long-context multilingual text encoder models, we implement several enhancements to BERT \cite{devlin-etal-2019-bert} architecture and train it from scratch using the vocabulary of XLM-R\footnote{
\url{https://hf.co/FacebookAI/xlm-roberta-base}
} \cite{conneau-etal-2020-unsupervised} series.

Specifically, we replace the absolute positional embeddings with RoPE  \cite{su2024roformer}, and upgrade the feedforward network (FFN) to gated linear unit (GLU) \cite{shazeer2020glu}.
To ensure compatibility with libraries like \texttt{FlashAttention} \citep{dao2023flashattention2}, we remove the dropout applied to attention scores.
In addition, we pad the token embedding size to be a multiple of 64, which could speedup the model throughput \cite{portes2023mosaicbert}.

\paragraph{Unpadding Mode}
Inspired by \citet{portes2023mosaicbert}, we unpad the input batch to reduce redundant computations associated with padding tokens (Figure \ref{fig:encoder}).
We use \texttt{xFormers} \citep{xFormers2022} to implement the variable length attention.
It dispatch the attention forward and backward to different kernels\footnote{
We adopt the memory-efficient attention \cite{rabe2021self} in this work.
} based on the numerical precision, attention head size and device type.
We unpad the MLM labels as well to reduce the computation cost of predicting non-masked tokens.

\paragraph{Data}
We assemble our multilingual pre-training data from a combination of the following sources:
C4 \citep{raffel2020exploring}, Skypile \citep{wei2023skywork} (2021-2023 subsets), mC4 \citep{xue-etal-2021-mt5}, CulturaX \citep{nguyen2023culturax}, Wikipedia \cite{wikidump} and books (proprietary).
We filter them and curate a dataset covering 75 Languages.
Appendix Table \ref{tab:mlm-data} presents the statistics of our dataset.

\paragraph{Training Curriculum}
We pre-train the model via masked language modeling (MLM) \citep{devlin-etal-2019-bert}\footnote{
We remove the next sentence prediction objective of BERT following \citep{liu2019roberta}.
}.
The MLM probability is set to $30\%$ \cite{portes2023mosaicbert}.
Following \citet{conneau2019cross} and \citet{conneau-etal-2020-unsupervised}, the data from different languages is sampled by a multinomial distribution with probabilities $\{q_i\}_{i=1 \dots N}$, where
\begin{equation}\label{eq:xlm-sampling}
q_i = \frac{p^\alpha_i}{\sum_{j=1}^N p^\alpha_j} \text{~~with~~} p_i = \frac{n_i}{\sum_{j=1}^N n_j} ,
\end{equation}
and $n_i$ is the number of texts in language $i$.
We set $\alpha=0.5$.
This sampling strategy could increase texts from low-resource languages.
To train the native 8192-context model more efficiently, we adopt a phased training curriculum \cite{xiong-etal-2024-effective}:
\begin{compactitem}
\item MLM-2048: we chunk the input into 2048 tokens and set RoPE base to $10,000$.
\item MLM-8192: we chunk the input into 8192 tokens and set RoPE base to $160,000$.
\end{compactitem}
Through this method, we could train the model with a large context length in limited resources
\footnote{
In our early experiments of English models, we investigated continue training by RetroMAE \citep{xiao-etal-2022-retromae} after MLM-8192.
However, we did not observe any improvement.
}.

\paragraph{Training Setup}
Following \citet{portes2023mosaicbert}, we use the learning rate decoupled\footnote{
However, \citet{xie2024overlooked} state that the decoupled weight decay is not ideal.
We recommend to keep the default setting.
} AdamW \citep{loshchilov2018decoupled} with weight decay $1e-5$.
We disable gradient clipping (set to 0) \citep{liu2019roberta}.
All models are trained on A100 GPU servers by BF16 PyTorch native automatic mixed precision via \texttt{transformers} \cite{wolf-etal-2020-transformers}.
We list the detailed hyper-parameters of each training stage in Appendix \ref{sec:app:mlm-training} and Table \ref{tab:hparam-mlm}.
We denote the resulting models as \texttt{\modelname-MLM-2048/8192}.

\begin{figure}
\includegraphics[width=\columnwidth]{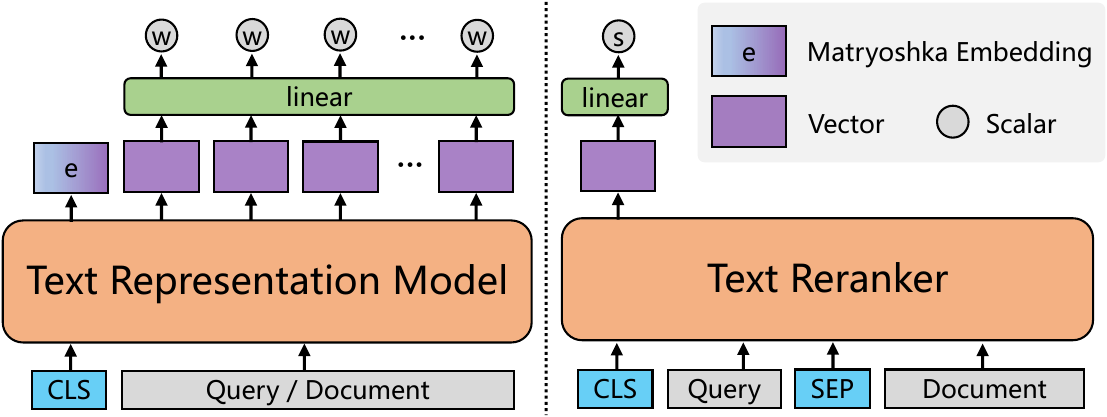}
\caption{
Our TRM and reranker.
}
\label{fig:trm-reranker}
\end{figure}

\subsection{Text Representation Model}\label{sec:method:representation}
Based on our encoder, we construct the TRM for the first-stage text retrieval in two steps: contrastive pre-training and fine-tuning \cite{wang2022text,li2023towards}.
Both steps share the same InfoNCE \cite{oord2018representation} learning objective:
\begin{equation} \label{eq:infonce}
\mathcal{L} = - \log \frac{ \text{exp}(s(q, d^+) / \tau) }{
\sum_{i=1}^{N}\text{exp}(s(q, d^i)/ \tau)
}  ,
\end{equation}
where $\tau$, $q$, and $d$ denote the temperature parameter, query and document.
The positive $d^+$ is the relevant document to $q$, and other irrelevant documents are negatives. These negatives can be either hard-negatives or in-batch negatives (documents of other instances in the same batch).
$s(q, d)$ is the relevance score of $q$ and $d$, measured by the dot product or cosine similarity between their respective representations.

\paragraph{Contrastive Pre-Training}
We take the encoder output hidden state of the [CLS] token as the dense representation (\ie embedding) and compute the relevance score by cosine similarity.
Our pre-training data (Appendix Table \ref{tab:unsupervised-data}) comprise naturally occurring text pairs (\eg question-answer pairs from Quora and StackExchange, title-content pairs of CommonCrawl), translation pairs \cite{nllb2024scaling}, and crosslingual instruction tuning data \cite{muennighoff2022crosslingual}.
We train the model with a batch size of $16,384$ and a learning rate of $5e-4$ for $240k$ steps.
Each batch is sampled from a single data source by the same distribution of Eq.\ref{eq:xlm-sampling}.
The queries (\resp documents) are truncated to the max tokens of $512$ (\resp $1024$).
We reverse scale the RoPE base from $160,000$ to $20,000$ to fit the $1024$ context length and acquire the long-context retrieval ability (denotes revNTK, ablation in \S\ref{sec:eval:analysis}).
We set $\tau$ of InfoNCE to $0.01$ and only use in-batch negatives.
More details refer to Appendix \ref{sec:app:trm-training}.
We denote this contrastive pre-trained model as \texttt{\modelname-CPT}, which is actually an unsupervised embedding model.

\paragraph{Matryoshka Embedding}
Many of recently released models and APIs offer elastic embeddings by Matryoshka representation learning (MRL) \cite{kusupati2022matryoshka}, providing competitive sub-vectors of embeddings to save index storage and speedup search.
Let $\bm{e} \in \mathbb{R}^{H}$ denotes an embedding and $\bm{e}_{:d}$ is the sliced sub-vector from dimension $0$ to $d < H$.
MRL\footnote{
Here we mean the MRL-E in \citet{kusupati2022matryoshka}.
} optimizes the weighted sum of multiple losses from different $d$ dimensional sub-vectors, \ie compute InfoNCE by $s_d(\bm{e}^q_{:d}, \bm{e}^d_{:d})$.
We add this objective to our TRM fine-tuning stage.

\paragraph{Sparse Representation}
\citet{chen-etal-2024-m3} show that neural sparse representations (term/token weights predicted by TRM) could greatly improve the long-context retrieval performance.
We follow this design, computing the term weight $w_t$ of each token of the input by $w_t = \text{ReLU}(\bm{W}\bm{h}_t)$, where $\bm{h}_t$ is the encoder hidden state of token $t$ with dimension size $H$ and $\bm{W} \in \mathbb{R}^{H \times 1}$ is randomly initialized.
If a token appears multiple times in the text, we keep the max weight.
The relevance score is computed by the joint importance of the co-occurring terms (denoted as \(q \cap d\)) within the query and document pair: \(s_{\text{sparse}}(q, d) = \sum_{t \in q \cap d} (w^q_t \cdot w^d_t)\). This is then used to derive the InfoNCE loss for training.

\begin{table*}
\centering
\setlength{\tabcolsep}{3pt}
\resizebox{\textwidth}{!}{
\begin{tabular}{lccccccccccc}
\toprule
\multirow{2}{*}{\bf Model} & \multirow{2}{*}{\bf Avg.} & Pair Class. & M.C. & \multicolumn{2}{c}{Structure Prediction} & \multicolumn{3}{c}{Question Answering} & \multicolumn{3}{c}{Cross-lingual Retrieval} \\
&& XNLI & XCOPA & UDPOS & WikiANN & XQuAD & MLQA  & TyDiQA-GoldP & Mewsli-X & LAReQA & Tatoeba\\
\midrule
\multicolumn{2}{c}{\#Languages (Total 50)} & 15 & 11 & 38 & 47 & 11 & 7 & 9 & 38 & 11 & 38 \\
\multicolumn{2}{c}{Metrics} & Acc. & Acc. & F1 & F1 & F1/EM & F1/EM & F1/EM & mAP@20 & mAP@20 & Acc. \\
\midrule
mBERT-base           & 59.43     & 66.63 & 55.49 & 71.80 & \bf 62.34 & 66.23 / 51.03 & 57.37 / 42.44 & \bf 55.01 / 38.05 & \bf 44.65 & \bf 75.26 & 39.49 \\
XLM-R-base           & 62.02     & \bf 74.50 & 50.45 & \bf 73.84 & 61.23 & 72.83 / 58.01 & 61.54 / 46.45 & 53.09 / 37.11 & 42.09 & 63.43 & 67.20 \\
\bf \modelname-MLM-2048    & \bf 65.24 & 73.17 & \bf 63.62 & 73.25 & 60.87 & \bf 75.33 / 60.00 & 64.02 / 48.57 & 53.58 / 36.68 & 44.41 & 72.13 & \bf 72.02 \\
\bf \modelname-MLM-8192    & 64.44     & 73.37 & 61.98 & 73.14 & 59.83 & 74.81 / 59.37 & \bf 64.24 / 48.80 & 49.85 / 33.27 & 44.52 & 71.54 & 71.10 \\
\bottomrule
\end{tabular}
}
\caption{
XTREME-R \cite{ruder-etal-2021-xtreme} results in the cross-lingual zero-shot transfer (models are trained on English data) setting.
M.C. stands for Multiple Choice.
The EM scores are not included in the average.
}
\label{tab:xtremer}
\end{table*}

\paragraph{Contrastive Fine-Tuning}
Now we construct the TRM by multi-task learning of matryoshka embedding and sparse representation:
\begin{equation}
\mathcal{L}_{\text{TRM}} = \lambda \mathcal{L}_{\text{sparse}} + \sum_{d \in D} w_d \mathcal{L}_{:d} \text{~},
\end{equation}
where $D = \{ 32k \mid k \in \mathbb{N}, k \geq 1, 32k \leq H \}$ is MRL dimension set, $w_d$ is the weight of dimension $d$, and $\lambda$ is the weight of sparse representation loss.
We fine-tune our contrastive pre-trained embedding model on diverse high-quality datasets with hard-negatives (\eg MS MARCO \cite{DBLP:conf/nips/NguyenRSGTMD16}, MIRACL \cite{10.1162/tacl_a_00595}, listed in Table \ref{tab:ft-data}).
We adopt a dynamic batching strategy \cite{chen-etal-2024-m3} to fine-tune 8192-context data.
The batch sampling strategy is the same as the pre-training stage.
The $\tau$ of MRL and sparse is set to $0.05$ and $0.01$ respectively.
Other details refer to Appendix \ref{sec:app:trm-training}.
We denote this fine-tuned model as \texttt{\modelname-TRM}.

\begin{table}
\centering
\resizebox{\columnwidth}{!}{
\begin{tabular}{lcccc}
\toprule
\textbf{Model} & \textbf{Params} & \bf Pos. & \bf Seq. Len. & \bf GLUE Avg. \\
\midrule
RoBERTa-base$^\alpha$         & 125M & Abs.  & 512  & 86.4 \\
\midrule
XLM-R-base           & 279M & Abs.  & 512  & 80.44 \\
\bf \modelname-MLM-2048    & \multirow{2}{*}{305M} & \multirow{2}{*}{RoPE}  & 2048 & 83.42 \\
\bf \modelname-MLM-8192    &  &  & 8192 & \bf 83.47 \\
\bottomrule
\end{tabular}
}
\caption{
GLUE \cite{wang2018glue} devset averages (w/o WNLI).
The detailed scores for each subset are shown in Table \ref{tab:glue-full}.
$^\alpha$Taken from Table 8 of \citet{liu2019roberta}.
The rest are from our runs, refer to Appendix \ref{sec:app:glue}.
}
\label{tab:glue-avg}
\end{table}

\subsection{Text Reranking Model}\label{sec:method:reranking}
We also build a reranker using the cross-encoder architecture.
It takes the query $q$ and document $d$ together as input: [CLS] $q$ [SEP] $d$, and directly predicts their relevance score by the [CLS] output state: $s_\text{rerank} = \bm{W} \bm{h}_\text{[CLS]} $.
In our experiment, $\bm{W} \in \mathbb{R}^{H \times 1}$ is randomly initialized.

The model is fine-tuned by InfoNCE in one step\footnote{
We found that the contrastive pre-training of reranker does not improve the performance.
} based on our pre-trained 8k-context text encoder model.
Unless otherwise specified, we employ identical data and training settings as our TRM fine-tuning stage (\S\ref{sec:method:representation}).
The difference lies in our adjustment of the hard-negatives.
We describe the detailed settings in Appendix \ref{sec:app:reranker-training}.
We denote this model as \texttt{\modelname-reranker}.

\begin{table}
\centering
\setlength{\tabcolsep}{3pt}
\resizebox{\columnwidth}{!}{
\begin{tabular}{ll|ccccc}
\toprule
\bf Model  & \bf Seq. & \bf en & \bf zh & \bf fr & \bf pl \\
\midrule
BGE-M3-unsupervised$^\dagger$ & 8192 & 56.48 & 57.53 & 57.95 & 55.98 \\
\multirow{2}{*}{\bf \modelname-CPT}
& 512$^*$ & \bf 60.16 & \bf 58.67 & 59.72 & \bf 57.66  \\
& 8192    & 60.04 & 58.63 & \bf 59.74 & 57.11  \\
\midrule
mE5-base           & 514  & 59.45 & 56.21 & 56.19 & 55.62 \\
mE5-large          & 514  & \bf 61.50  & 58.81 & 56.07 & 60.08 \\
BGE-M3 (Dense)$^\dagger$ & 8192 & 59.84 & 60.80 & 58.79 & \bf 60.35 \\
\textbf{\modelname-TRM} (Dense) & 8192 & 61.40 & \bf 62.72 & \bf 59.79 & 58.22 \\
\midrule
E5-mistral-7b    & 32768 & 66.63 & 60.81 & 48.33 & - \\
voyage-multilingual-2 & 32000 & - & - & 61.65 & - \\
Cohere-multilingual-v3.0 & 512 & 64.01 & - & 56.02 & - \\
OpenAI-3-large & 8191 & 64.59 & - & - & - \\
OpenAI-3-small & 8191 & 62.26 & - & - & - \\
\bottomrule
\end{tabular}
}
\caption{
Embedding model performance on MTEB English \citep{muennighoff-etal-2023-mteb}, Chinese \citep{xiao2023c}, French \cite{ciancone2024mteb-french} and Polish \citep{poswiata2024pl}.
The scores of other models are retrieved from the MTEB online leaderboard.
$^*$To be consistent with the setting in contrastive pre-training, in retrieval tasks, the max sequence length of the document side is set to 1024.
$^\dagger$Denote our runs.
}
\label{tab:mteb-multi}
\end{table}

\begin{table*}
\centering
\resizebox{\textwidth}{!}{
\begin{tabular}{lcc|c|ccccc}
\toprule
 & Params & Seq. Len. & Avg. & MLDR & MIRACL & MKQA & BEIR & LoCo \\
Metric & & & & nDCG@10 & nDCG@10 & recall@20 & nDCG@10 & nDCG@10 \\
\#languages (Total 33) &&&& 13 & 18 & 25 & 1 & 1 \\
\midrule
BM25 & - & - & 47.0 & 53.6 & 31.9 & 28.1 & 41.7 & 79.9 \\
mE5-base & 279M & 514 & 53.5 & 30.5 & 62.3 & 53.7 & 48.9 & 72.2 \\
mE5-large & 560M & 514 & 57.7 & 34.2 & 65.4 & 63.5 & 51.4 & 74.3 \\
E5-mistral-7b & 7111M & 32768 & 62.4 & 42.6 & 62.2 & 62.4 & 56.9 & 87.8 \\
OpenAI-3-large & - & 8191 & - & - & 54.9 & 62.1 & 55.4 & 79.4 \\
\midrule
BGE-M3 Dense & \multirow{3}{*}{568M} &  \multirow{3}{*}{8192} & 64.3 & 52.5 & 67.7 & 67.8 & 48.7 & 84.9 \\
BGE-M3 Sparse & & & 55.1 & 62.2 & 53.9 & 36.3 & 38.3 & 84.9 \\
BGE-M3 Dense + Sparse & & & 67.7 & 64.8 & \textbf{68.9} & \textbf{68.1} & 49.4 & 87.4 \\
\midrule
\textbf{\modelname-TRM} Dense & \multirow{3}{*}{304M} &  \multirow{3}{*}{8192} & 66.7 & 56.6 & 62.1 & 65.8 & 51.1 & 88.9 \\
\textbf{\modelname-TRM} Sparse & & & 57.2 & 71.0 & 55.9 & 31.6 & 39.2 & 88.1 \\
\textbf{\modelname-TRM} Dense + Sparse & & & \textbf{68.9} & \textbf{71.3} & 64.5 & 66.0 & \textbf{51.4} & \textbf{91.3} \\
\bottomrule
\end{tabular}
}
\caption{
Retrieval results on MIRACL \cite{10.1162/tacl_a_00595} and MLDR \cite{chen-etal-2024-m3} (multilingual), MKQA \cite{longpre2021mkqa} (crosslingual), BEIR \cite{thakur2beir} and LoCo \cite{saadbenchmarking} (English).
}\label{tab:retrieval-avg}
\end{table*}

\section{Evaluation}\label{sec:eval}
We separately evaluate our text encoder in \S\ref{sec:eval:encoder}, TRM and reranker in \S\ref{sec:eval:representation} and \S\ref{sec:eval:multi-stage-retrieval}.

\subsection{Natural Language Understanding}\label{sec:eval:encoder}
We evaluate the encoder on the cross-lingual natural language understanding (NLU) benchmark XTREME-R\footnote{
We use XTREME-R \cite{ruder-etal-2021-xtreme} instead of XTREME \cite{hu2020xtreme} since we found the retrieval tasks of XTREME is unstable and difficult to evaluate.
} \cite{ruder-etal-2021-xtreme} and the English NLU benchmark  GLUE \cite{wang2018glue}.
Results show that our encoder outperforms the same-sized previous state-of-the-art XLM-R \cite{conneau-etal-2020-unsupervised} on all benchmarks.

\paragraph{XTREME-R}
We focus on the \textit{zero-shot cross-lingual transfer} setting where models are fine-tuned on English trainset and tested on multi- and cross-lingual data.
The fine-tuning setup is described in Appendix \ref{sec:app:xtremer}.
We run mBERT-base, XLM-R-base, and our encoder, as shown in Table \ref{tab:xtremer}.
Our 2048 and 8192 encoder models achieve average scores that are higher than those of XLM-R by $3.22$ and $2.42$ points, respectively.

\paragraph{GLUE}
We also report the performance on the devset of GLUE benchmark \cite{wang2018glue}.
The fine-tuning details refer to Appendix \ref{sec:app:glue}.
Table \ref{tab:glue-avg} presents the average scores (Table \ref{tab:glue-full} provides the full results).
Our models consistently outperform XLM-R-base and reasonably lag behind the English RoBERTa-base \cite{liu2019roberta}.

\begin{figure}
\includegraphics[width=\columnwidth]{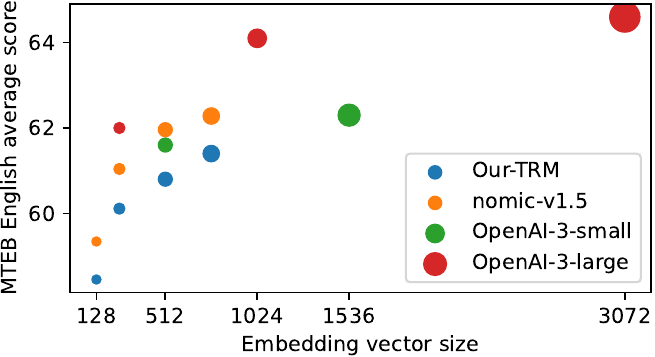}
\caption{
Elastic embedding results on MTEB English.
}
\label{fig:mrl-mteb}
\end{figure}

\subsection{Text Embedding}\label{sec:eval:representation}
Our contrastive pre-training actually yields a text embedding model.
To understand the pre-training and fine-tuning of TRM, and to compare with other models, we first run the most popular text embedding benchmark MTEB \cite{muennighoff-etal-2023-mteb} as well as its Chinese, French and Polish versions.

\paragraph{Multilingual MTEB}
The results in Table \ref{tab:mteb-multi} 
also present the scores of LLM-based models and commercial APIs for reference.
For contrastive pre-trained models, our model outperforms \texttt{BGE-M3-unsupervised} \cite{chen-etal-2024-m3} on all four subsets, through our backbone has fewer params than \texttt{XLM-R-large}.
Comparing with \texttt{BGE-M3} and \texttt{mE5} \cite{wang2024multilingual}, our final TRM achieves best scores on Chinese and French, and is competitive on English.

\paragraph{Elastic Embedding}
We compare our TRM (only elastic embeddings) with open-source model and commercial APIs on MTEB English (Figure \ref{fig:mrl-mteb}).
Our model presents close scores to the same-sized English-only \texttt{nomic-v1.5}, which is promising for a multilingual model.
However, it is still behind OpenAI APIs, which is reasonable since they are guessed to be much larger models.

\begin{table*}
\centering
\resizebox{\textwidth}{!}{
\begin{tabular}{lcc|c|cccc}
\toprule
 & Params & Seq. Len. & Avg. & MLDR & MIRACL & MKQA & BEIR \\
Metric & & & & nDCG@10 & nDCG@10 & recall@20 & nDCG@10  \\
\#languages (Total 33) &&&& 13 & 18 & 25 & 1 \\
\midrule
Retrieval (\modelname-TRM Dense) & 304M & 8192 & 58.9 & 56.6 & 62.1 & 65.8 & 50.9 \\
\midrule
jina-reranker-v2-multilingual & 278M	& 8192 & 59.4 & 53.2 & 65.8 & \bf 68.8 & 49.7    \\
bge-reranker-v2-m3 & 568M & 8192 & 65.7 & 66.8 &  \bf 72.6  &   68.7  & 54.6 \\
\bf \modelname-reranker & 304M & 8192 & \bf 67.4 & \bf 78.7    &  68.5      &  67.2    & \bf 55.4     \\
\bottomrule
\end{tabular}
}
\caption{
Results of reranking based on the candidates retrieved by our TRM dense model (refer to Table \ref{tab:retrieval-avg}).
}
\label{tab:rerank}
\end{table*}

\subsection{Text Retrieval}\label{sec:eval:multi-stage-retrieval}
We conduct evaluations to our TRM and reranker on retrieval benchmarks in multilingual (Miracl \cite{10.1162/tacl_a_00595} and MLDR \cite{chen-etal-2024-m3}), crosslingual (MKQA \cite{longpre2021mkqa}) setting, and the commonly used English BEIR \cite{thakur2beir} and LoCo \cite{saadbenchmarking}.
Our models are close to the state-of-the-art large models on Miracl, MKQA and BEIR, while achieve better scores on long-context datasets MLDR and LoCo.
Details are in Appendix \ref{sec:app:retrieval}.

\paragraph{First-Stage Retrieval}
We compare our TRM to the hybrid model \texttt{BGE-M3} \cite{chen-etal-2024-m3}, dense models like \texttt{mE5} \cite{wang2024multilingual} and \texttt{E5-mistral-7b} \cite{wang2024improving}, and BM25.
As shown in Table \ref{tab:retrieval-avg}, our TRM consistently outperforms \texttt{mE5} and OpenAI APIs, better than \texttt{BGE-M3} on MLDR, and close to it on the rest parts.

\paragraph{Reranking}
In Table \ref{tab:rerank}, we evaluate rerankers based on the candidates retrieved by Our-TRM dense model.
Our model outperforms the powerful \texttt{bge-reranker-v2-m3} \cite{chen-etal-2024-m3} with a smaller size.
Moreover, it greatly surpasses the same-sized \texttt{jina-reranker-v2-multilingual}.

\begin{table}
\centering
\setlength{\tabcolsep}{3pt}
\resizebox{\columnwidth}{!}{
\begin{tabular}{lcccccc}
\toprule
\bf Model  & \bf Attn. & \bf Unpad. & \bf \begin{tabular}{c}Encoding\\Time \end{tabular} & \bf \begin{tabular}{c}Search\\Latency \end{tabular} \\
\midrule
\multirow{2}{*}{BGE-M3} & eager & \multirow{2}{*}{$\times$} & 1800s & \multirow{2}{*}{20.35ms} \\
& SDPA-MEA &  & 744s \\
\midrule
\multirow{5}{*}{\bf\modelname-TRM} & eager & $\times$ & 695s & \multirow{5}{*}{15.07ms} \\
& SDPA-MEA & $\times$ & 298s \\
& eager & \checkmark & 675s \\
& SDPA-MEA & \checkmark & 279s \\
& MEA & \checkmark & 52s \\
\bottomrule
\end{tabular}
}
\caption{
Dense retrieval efficiency.
Encoding time is running MLDR-hi corpus (3806 texts with average 4456 tokens after truncating to maximum 8192) on one A100 GPU with FP16.
Search latency is measured on a faiss index with 8.8M texts.
MEA is the memory-efficient attention in \texttt{xFormers}.
SDPA-MEA denotes MEA dispatched by scaled dot-product attention of \texttt{PyTorch}.
}
\label{tab:efficiency}
\end{table}

\subsection{Analysis}\label{sec:eval:analysis}

\paragraph{Efficiency}
We compare the efficiency of our TRM with \texttt{BGE-M3} on dense retrieval in Table \ref{tab:efficiency}.
To simulate the real-world scenario, the encoding time is the duration of encoding texts without length grouping.
Our TRM is up to 14 times faster than \texttt{BGE-M3} (52s \emph{v.s.} 744s).
The end-to-end unpadding with \texttt{xFormers} is crucial for encoding, which reduces the time by 5 times (52s \emph{v.s.} 279s).

\begin{figure}
\includegraphics[width=\columnwidth]{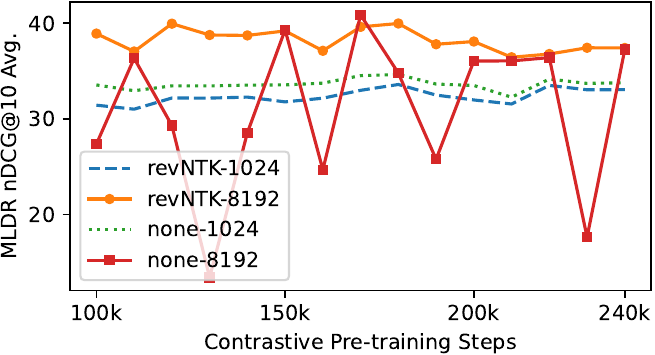}
\caption{
MLDR scores in contrastive pre-training.
\texttt{none} keeps the RoPE untouched in pre-training.
1024 and 8192 are the max sequence length in evaluations.
\texttt{revNTK-8912} recovers the 8k context by NTK scaling.
}
\label{fig:rope-cpt}
\end{figure}

\paragraph{Scaled Contrastive Pre-Training}
We utilize the reversed NTK scaling in contrastive pre-training to reduce required text length, where we set the RoPE base to $1/8$ of the original and train the 8k encoder with 1k max length.
To evaluate the effectiveness, we run the same training without the reversed NTK, comparing the MLDR scores in Figure \ref{fig:rope-cpt}.
With revNTK, models exhibit slightly lower performance on 1k context but achieve more stable 8k performance across different training steps.

\section{Related Work}
\label{sec:related}

Training long-context TRMs has become a hot topic recently.
OpenAI released 8191 context APIs \cite{neelakantan2022text} have set the target for open-source community.
\citet{portes2023mosaicbert} and \citet{gunther2023jina2} replace position embedding of BERT with Alibi \cite{press2022train} attention bias and pre-train from scratch, which is shown to be effective in build 8k TRMs.
\citet{nussbaum2024nomic} explore the more powerful RoPE \cite{su2024roformer} in BERT pre-training and their 2048-context pre-trained encoder achieve better retrieval performance on English.
\citet{zhu2024longembed} suggest patch E5 \cite{wang2022text} with RoPE.
We also use RoPE and provide multi-stage training for native 8192-context text encoder, TRM, and reranker.

\citet{chen-etal-2024-m3} propose long-context multilingual TRM and reranker based on XLM-RoBERTa-large \cite{conneau-etal-2020-unsupervised} by extending position embedding to 8192 via continue training.
We pre-train native 8k multilingual models from scratch for better long-context performance and efficiency.

\section{Conclusion}
We present the holistic practice of building native 8192-context multilingual retrieval models.
We first suggest a text encoder with RoPE and unpadding, which is pre-trained by a two-stage MLM curriculum for 8k context.
Evaluations on NLU benchmarks show that our encoder outperforms XLM-RoBERTa in the same size.
Based on our encoder, we construct a hybrid TRM and a cross-encoder reranker by contrastive learning.
The TRM is pre-trained with reversed RoPE NTK scaling 
and fine-tuned to generate both Matryoshka embeddings and sparse representations.
Results on monolingual and crosslingual retrieval benchmarks show that our TRM and reranker are close to larger ones on regular datasets, and achieve better performance on long-context datasets.
This means our models are more efficient for industrial applications.

\section*{Acknowledgements}
This work was supported by Alibaba Research Intern Program.

\bibliography{custom}

\begin{thebibliography}{90}
\providecommand{\natexlab}[1]{#1}

\bibitem[{Artetxe et~al.(2020)Artetxe, Ruder, and Yogatama}]{artetxe-etal-2020-cross}
Mikel Artetxe, Sebastian Ruder, and Dani Yogatama. 2020.
\newblock \href {https://doi.org/10.18653/v1/2020.acl-main.421} {On the cross-lingual transferability of monolingual representations}.
\newblock In \emph{Proc. of the ACL}, pages 4623--4637, Online.

\bibitem[{Bonifacio et~al.(2021)Bonifacio, Jeronymo, Abonizio, Campiotti, Fadaee, Lotufo, and Nogueira}]{bonifacio2021mmarco}
Luiz Bonifacio, Vitor Jeronymo, Hugo~Queiroz Abonizio, Israel Campiotti, Marzieh Fadaee, Roberto Lotufo, and Rodrigo Nogueira. 2021.
\newblock \href {https://arxiv.org/abs/2108.13897} {{mMARCO}: A multilingual version of the ms marco passage ranking dataset}.
\newblock \emph{arXiv preprint arXiv:2108.13897}.

\bibitem[{Cer et~al.(2017)Cer, Diab, Agirre, Lopez-Gazpio, and Specia}]{cer-etal-2017-semeval}
Daniel Cer, Mona Diab, Eneko Agirre, I{\~n}igo Lopez-Gazpio, and Lucia Specia. 2017.
\newblock \href {https://doi.org/10.18653/v1/S17-2001} {{S}em{E}val-2017 task 1: Semantic textual similarity multilingual and crosslingual focused evaluation}.
\newblock In \emph{Proc. of the 11th International Workshop on Semantic Evaluation ({S}em{E}val-2017)}, pages 1--14, Vancouver, Canada.

\bibitem[{Chen et~al.(2024)Chen, Xiao, Zhang, Luo, Lian, and Liu}]{chen-etal-2024-m3}
Jianlyu Chen, Shitao Xiao, Peitian Zhang, Kun Luo, Defu Lian, and Zheng Liu. 2024.
\newblock \href {https://doi.org/10.18653/v1/2024.findings-acl.137} {{M}3-embedding: Multi-linguality, multi-functionality, multi-granularity text embeddings through self-knowledge distillation}.
\newblock In \emph{Findings of the ACL}, pages 2318--2335, Bangkok, Thailand and virtual meeting.

\bibitem[{Chen et~al.(2016)Chen, Xu, Zhang, and Guestrin}]{chen2016training}
Tianqi Chen, Bing Xu, Chiyuan Zhang, and Carlos Guestrin. 2016.
\newblock \href {https://arxiv.org/abs/1604.06174} {Training deep nets with sublinear memory cost}.
\newblock \emph{arXiv preprint arXiv:1604.06174}.

\bibitem[{Ciancone et~al.(2024)Ciancone, Kerboua, Schaeffer, and Siblini}]{ciancone2024mteb-french}
Mathieu Ciancone, Imene Kerboua, Marion Schaeffer, and Wissam Siblini. 2024.
\newblock \href {https://arxiv.org/abs/2405.20468} {{MTEB-F}rench: Resources for french sentence embedding evaluation and analysis}.
\newblock \emph{arXiv preprint arXiv:2405.20468}.

\bibitem[{Clark et~al.(2020)Clark, Choi, Collins, Garrette, Kwiatkowski, Nikolaev, and Palomaki}]{clark-etal-2020-tydi}
Jonathan~H. Clark, Eunsol Choi, Michael Collins, Dan Garrette, Tom Kwiatkowski, Vitaly Nikolaev, and Jennimaria Palomaki. 2020.
\newblock \href {https://doi.org/10.1162/tacl_a_00317} {{T}y{D}i {QA}: A benchmark for information-seeking question answering in typologically diverse languages}.
\newblock \emph{Transactions of the Association for Computational Linguistics}, 8:454--470.

\bibitem[{Conneau et~al.(2020)Conneau, Khandelwal, Goyal, Chaudhary, Wenzek, Guzm{\'a}n, Grave, Ott, Zettlemoyer, and Stoyanov}]{conneau-etal-2020-unsupervised}
Alexis Conneau, Kartikay Khandelwal, Naman Goyal, Vishrav Chaudhary, Guillaume Wenzek, Francisco Guzm{\'a}n, Edouard Grave, Myle Ott, Luke Zettlemoyer, and Veselin Stoyanov. 2020.
\newblock \href {https://doi.org/10.18653/v1/2020.acl-main.747} {Unsupervised cross-lingual representation learning at scale}.
\newblock In \emph{Proc. of the ACL}, pages 8440--8451, Online.

\bibitem[{Conneau and Lample(2019)}]{conneau2019cross}
Alexis Conneau and Guillaume Lample. 2019.
\newblock \href {https://proceedings.neurips.cc/paper/2019/hash/c04c19c2c2474dbf5f7ac4372c5b9af1-Abstract.html} {Cross-lingual language model pretraining}.
\newblock In \emph{Proc. of the 33rd NeurIPS}, pages 7057--7067.

\bibitem[{Conneau et~al.(2018)Conneau, Rinott, Lample, Williams, Bowman, Schwenk, and Stoyanov}]{conneau-etal-2018-xnli}
Alexis Conneau, Ruty Rinott, Guillaume Lample, Adina Williams, Samuel Bowman, Holger Schwenk, and Veselin Stoyanov. 2018.
\newblock \href {https://doi.org/10.18653/v1/D18-1269} {{XNLI}: Evaluating cross-lingual sentence representations}.
\newblock In \emph{Proc. of the EMNLP}, pages 2475--2485, Brussels, Belgium.

\bibitem[{Dadas et~al.(2024)Dadas, Pere{\l}kiewicz, and Po{\'s}wiata}]{dadas2024pirb}
Slawomir Dadas, Micha{\l} Pere{\l}kiewicz, and Rafa{\l} Po{\'s}wiata. 2024.
\newblock \href {https://aclanthology.org/2024.lrec-main.1117} {{PIRB}: A comprehensive benchmark of {P}olish dense and hybrid text retrieval methods}.
\newblock In \emph{Proc. of the LREC-COLING}, pages 12761--12774, Torino, Italia. ELRA and ICCL.

\bibitem[{Dao(2023)}]{dao2023flashattention2}
Tri Dao. 2023.
\newblock \href {https://openreview.net/forum?id=mZn2Xyh9Ec} {Flashattention-2: Faster attention with better parallelism and work partitioning}.
\newblock In \emph{The Twelfth International Conference on Learning Representations}.

\bibitem[{de~Marneffe et~al.(2021)de~Marneffe, Manning, Nivre, and Zeman}]{de2021universal}
Marie-Catherine de~Marneffe, Christopher~D. Manning, Joakim Nivre, and Daniel Zeman. 2021.
\newblock \href {https://doi.org/10.1162/coli_a_00402} {{Universal Dependencies}}.
\newblock \emph{Computational Linguistics}, 47(2):255--308.

\bibitem[{Devlin et~al.(2019)Devlin, Chang, Lee, and Toutanova}]{devlin-etal-2019-bert}
Jacob Devlin, Ming-Wei Chang, Kenton Lee, and Kristina Toutanova. 2019.
\newblock \href {https://doi.org/10.18653/v1/N19-1423} {{BERT}: Pre-training of deep bidirectional transformers for language understanding}.
\newblock In \emph{Proc. of the NAACL-HTL}, pages 4171--4186, Minneapolis, Minnesota.

\bibitem[{Dolan and Brockett(2005)}]{dolan2005automatically}
William~B. Dolan and Chris Brockett. 2005.
\newblock \href {https://aclanthology.org/I05-5002} {Automatically constructing a corpus of sentential paraphrases}.
\newblock In \emph{Proc. of the Third International Workshop on Paraphrasing ({IWP}2005)}.

\bibitem[{Facebook(2019)}]{tatoebav1}
Facebook. 2019.
\newblock Tatoeba test set.
\newblock \url{https://github.com/facebookresearch/LASER/tree/main/data/tatoeba/v1}.

\bibitem[{Foundation()}]{wikidump}
Wikimedia Foundation.
\newblock \href {https://dumps.wikimedia.org} {Wikimedia downloads}.

\bibitem[{Gao et~al.(2021{\natexlab{a}})Gao, Dai, and Callan}]{gao2021rethink}
Luyu Gao, Zhuyun Dai, and Jamie Callan. 2021{\natexlab{a}}.
\newblock \href {https://doi.org/10.1007/978-3-030-72240-1_26} {Rethink training of bert rerankers in multi-stage retrieval pipeline}.
\newblock In \emph{Proc. of the 43rd European Conference on IR Research}, page 280–286, Berlin, Heidelberg.

\bibitem[{Gao et~al.(2021{\natexlab{b}})Gao, Yao, and Chen}]{gao-etal-2021-simcse}
Tianyu Gao, Xingcheng Yao, and Danqi Chen. 2021{\natexlab{b}}.
\newblock \href {https://doi.org/10.18653/v1/2021.emnlp-main.552} {{S}im{CSE}: Simple contrastive learning of sentence embeddings}.
\newblock In \emph{Proc. of the EMNLP}, pages 6894--6910, Online and Punta Cana, Dominican Republic.

\bibitem[{Gao et~al.(2023)Gao, Xiong, Gao, Jia, Pan, Bi, Dai, Sun, and Wang}]{gao2023retrieval}
Yunfan Gao, Yun Xiong, Xinyu Gao, Kangxiang Jia, Jinliu Pan, Yuxi Bi, Yi~Dai, Jiawei Sun, and Haofen Wang. 2023.
\newblock \href {https://arxiv.org/abs/2312.10997} {Retrieval-augmented generation for large language models: A survey}.
\newblock \emph{arXiv preprint arXiv:2312.10997}.

\bibitem[{G{\"u}nther et~al.(2023)G{\"u}nther, Ong, Mohr, Abdessalem, Abel, Akram, Guzman, Mastrapas, Sturua, Wang et~al.}]{gunther2023jina2}
Michael G{\"u}nther, Jackmin Ong, Isabelle Mohr, Alaeddine Abdessalem, Tanguy Abel, Mohammad~Kalim Akram, Susana Guzman, Georgios Mastrapas, Saba Sturua, Bo~Wang, et~al. 2023.
\newblock \href {https://arxiv.org/abs/2310.19923} {Jina embeddings 2: 8192-token general-purpose text embeddings for long documents}.
\newblock \emph{arXiv preprint arXiv:2310.19923}.

\bibitem[{Hu et~al.(2020)Hu, Ruder, Siddhant, Neubig, Firat, and Johnson}]{hu2020xtreme}
Junjie Hu, Sebastian Ruder, Aditya Siddhant, Graham Neubig, Orhan Firat, and Melvin Johnson. 2020.
\newblock \href {http://proceedings.mlr.press/v119/hu20b/hu20b.pdf} {{XTREME}: A massively multilingual multi-task benchmark for evaluating cross-lingual generalisation}.
\newblock In \emph{International Conference on Machine Learning}, pages 4411--4421. PMLR.

\bibitem[{Huang et~al.(2024)Huang, Hu, Jing, Gao, and Wu}]{huang2024piccolo2}
Junqin Huang, Zhongjie Hu, Zihao Jing, Mengya Gao, and Yichao Wu. 2024.
\newblock \href {https://arxiv.org/abs/2405.06932} {Piccolo2: General text embedding with multi-task hybrid loss training}.
\newblock \emph{arXiv preprint arXiv:2405.06932}.

\bibitem[{Joshi et~al.(2017)Joshi, Choi, Weld, and Zettlemoyer}]{joshi-etal-2017-triviaqa}
Mandar Joshi, Eunsol Choi, Daniel Weld, and Luke Zettlemoyer. 2017.
\newblock \href {https://doi.org/10.18653/v1/P17-1147} {{T}rivia{QA}: A large scale distantly supervised challenge dataset for reading comprehension}.
\newblock In \emph{Proc. of the ACL}, pages 1601--1611, Vancouver, Canada.

\bibitem[{Kusupati et~al.(2022)Kusupati, Bhatt, Rege, Wallingford, Sinha, Ramanujan, Howard-Snyder, Chen, Kakade, Jain et~al.}]{kusupati2022matryoshka}
Aditya Kusupati, Gantavya Bhatt, Aniket Rege, Matthew Wallingford, Aditya Sinha, Vivek Ramanujan, William Howard-Snyder, Kaifeng Chen, Sham Kakade, Prateek Jain, et~al. 2022.
\newblock \href {https://proceedings.neurips.cc/paper_files/paper/2022/hash/c32319f4868da7613d78af9993100e42-Abstract-Conference.html} {Matryoshka representation learning}.
\newblock In \emph{Proc. of the 36th NeurIPS}, pages 30233--30249.

\bibitem[{Kwiatkowski et~al.(2019)Kwiatkowski, Palomaki, Redfield, Collins, Parikh, Alberti, Epstein, Polosukhin, Devlin, Lee, Toutanova, Jones, Kelcey, Chang, Dai, Uszkoreit, Le, and Petrov}]{kwiatkowski-etal-2019-natural}
Tom Kwiatkowski, Jennimaria Palomaki, Olivia Redfield, Michael Collins, Ankur Parikh, Chris Alberti, Danielle Epstein, Illia Polosukhin, Jacob Devlin, Kenton Lee, Kristina Toutanova, Llion Jones, Matthew Kelcey, Ming-Wei Chang, Andrew~M. Dai, Jakob Uszkoreit, Quoc Le, and Slav Petrov. 2019.
\newblock \href {https://doi.org/10.1162/tacl_a_00276} {Natural questions: A benchmark for question answering research}.
\newblock \emph{Transactions of the Association for Computational Linguistics}, 7:452--466.

\bibitem[{Lee et~al.(2024{\natexlab{a}})Lee, Roy, Xu, Raiman, Shoeybi, Catanzaro, and Ping}]{lee2024nv}
Chankyu Lee, Rajarshi Roy, Mengyao Xu, Jonathan Raiman, Mohammad Shoeybi, Bryan Catanzaro, and Wei Ping. 2024{\natexlab{a}}.
\newblock \href {https://arxiv.org/abs/2405.17428} {Nv-embed: Improved techniques for training llms as generalist embedding models}.
\newblock \emph{arXiv preprint arXiv:2405.17428}.

\bibitem[{Lee et~al.(2024{\natexlab{b}})Lee, Dai, Ren, Chen, Cer, Cole, Hui, Boratko, Kapadia, Ding et~al.}]{lee2024gecko}
Jinhyuk Lee, Zhuyun Dai, Xiaoqi Ren, Blair Chen, Daniel Cer, Jeremy~R Cole, Kai Hui, Michael Boratko, Rajvi Kapadia, Wen Ding, et~al. 2024{\natexlab{b}}.
\newblock \href {https://arxiv.org/abs/2403.20327} {Gecko: Versatile text embeddings distilled from large language models}.
\newblock \emph{arXiv preprint arXiv:2403.20327}.

\bibitem[{Lee et~al.(2024{\natexlab{c}})Lee, Shakir, Koenig, and Lipp}]{emb2024mxbai}
Sean Lee, Aamir Shakir, Darius Koenig, and Julius Lipp. 2024{\natexlab{c}}.
\newblock \href {https://www.mixedbread.ai/blog/mxbai-embed-large-v1} {Open source strikes bread - new fluffy embeddings model}.

\bibitem[{Lefaudeux et~al.(2022)Lefaudeux, Massa, Liskovich, Xiong, Caggiano, Naren, Xu, Hu, Tintore, Zhang, Labatut, and Haziza}]{xFormers2022}
Benjamin Lefaudeux, Francisco Massa, Diana Liskovich, Wenhan Xiong, Vittorio Caggiano, Sean Naren, Min Xu, Jieru Hu, Marta Tintore, Susan Zhang, Patrick Labatut, and Daniel Haziza. 2022.
\newblock xformers: A modular and hackable transformer modelling library.
\newblock \url{https://github.com/facebookresearch/xformers}.

\bibitem[{Lewis et~al.(2020)Lewis, Oguz, Rinott, Riedel, and Schwenk}]{lewis-etal-2020-mlqa}
Patrick Lewis, Barlas Oguz, Ruty Rinott, Sebastian Riedel, and Holger Schwenk. 2020.
\newblock \href {https://doi.org/10.18653/v1/2020.acl-main.653} {{MLQA}: Evaluating cross-lingual extractive question answering}.
\newblock In \emph{Proc. of ACL}, pages 7315--7330, Online.

\bibitem[{Li et~al.(2023)Li, Zhang, Zhang, Long, Xie, and Zhang}]{li2023towards}
Zehan Li, Xin Zhang, Yanzhao Zhang, Dingkun Long, Pengjun Xie, and Meishan Zhang. 2023.
\newblock \href {https://arxiv.org/abs/2308.03281} {Towards general text embeddings with multi-stage contrastive learning}.
\newblock \emph{arXiv preprint arXiv:2308.03281}.

\bibitem[{Liu et~al.(2024)Liu, Yan, An, Qiu, and Lin}]{liuscaling}
Xiaoran Liu, Hang Yan, Chenxin An, Xipeng Qiu, and Dahua Lin. 2024.
\newblock \href {https://openreview.net/forum?id=JO7k0SJ5V6} {Scaling laws of rope-based extrapolation}.
\newblock In \emph{The Twelfth International Conference on Learning Representations}.

\bibitem[{Liu et~al.(2019)Liu, Ott, Goyal, Du, Joshi, Chen, Levy, Lewis, Zettlemoyer, and Stoyanov}]{liu2019roberta}
Yinhan Liu, Myle Ott, Naman Goyal, Jingfei Du, Mandar Joshi, Danqi Chen, Omer Levy, Mike Lewis, Luke Zettlemoyer, and Veselin Stoyanov. 2019.
\newblock \href {https://arxiv.org/abs/1907.11692} {Roberta: A robustly optimized bert pretraining approach}.
\newblock \emph{arXiv preprint arXiv:1907.11692}.

\bibitem[{Long et~al.(2022)Long, Gao, Zou, Xu, Xie, Guo, Xu, Jiang, Xing, and Yang}]{long2022multi}
Dingkun Long, Qiong Gao, Kuan Zou, Guangwei Xu, Pengjun Xie, Ruijie Guo, Jian Xu, Guanjun Jiang, Luxi Xing, and Ping Yang. 2022.
\newblock \href {https://doi.org/10.1145/3477495.3531736} {Multi-cpr: A multi domain chinese dataset for passage retrieval}.
\newblock In \emph{Proc. of the 45th SIGIR}, page 3046–3056, New York, NY, USA. Association for Computing Machinery.

\bibitem[{Longpre et~al.(2021)Longpre, Lu, and Daiber}]{longpre2021mkqa}
Shayne Longpre, Yi~Lu, and Joachim Daiber. 2021.
\newblock \href {https://doi.org/10.1162/tacl_a_00433} {{MKQA}: A linguistically diverse benchmark for multilingual open domain question answering}.
\newblock \emph{Transactions of the Association for Computational Linguistics}, 9:1389--1406.

\bibitem[{Loshchilov and Hutter(2018)}]{loshchilov2018decoupled}
Ilya Loshchilov and Frank Hutter. 2018.
\newblock \href {https://openreview.net/forum?id=Bkg6RiCqY7} {Decoupled weight decay regularization}.
\newblock In \emph{International Conference on Learning Representations}.

\bibitem[{Manning(2008)}]{manning2008introduction}
Christopher~D Manning. 2008.
\newblock \emph{Introduction to information retrieval}.
\newblock Syngress Publishing,.

\bibitem[{Men et~al.(2024)Men, Xu, Wang, Zhang, Lin, Han, and Chen}]{men2024base}
Xin Men, Mingyu Xu, Bingning Wang, Qingyu Zhang, Hongyu Lin, Xianpei Han, and Weipeng Chen. 2024.
\newblock \href {https://arxiv.org/abs/2405.14591} {Base of rope bounds context length}.
\newblock \emph{arXiv preprint arXiv:2405.14591}.

\bibitem[{Muennighoff et~al.(2024)Muennighoff, Hongjin, Wang, Yang, Wei, Yu, Singh, and Kiela}]{muennighoff2024generative}
Niklas Muennighoff, SU~Hongjin, Liang Wang, Nan Yang, Furu Wei, Tao Yu, Amanpreet Singh, and Douwe Kiela. 2024.
\newblock \href {https://openreview.net/forum?id=8cQrRO9iFe} {Generative representational instruction tuning}.
\newblock In \emph{ICLR 2024 Workshop: How Far Are We From AGI}.

\bibitem[{Muennighoff et~al.(2023{\natexlab{a}})Muennighoff, Tazi, Magne, and Reimers}]{muennighoff-etal-2023-mteb}
Niklas Muennighoff, Nouamane Tazi, Loic Magne, and Nils Reimers. 2023{\natexlab{a}}.
\newblock \href {https://doi.org/10.18653/v1/2023.eacl-main.148} {{MTEB}: Massive text embedding benchmark}.
\newblock In \emph{Proc. of the EACL}, pages 2014--2037, Dubrovnik, Croatia.

\bibitem[{Muennighoff et~al.(2023{\natexlab{b}})Muennighoff, Wang, Sutawika, Roberts, Biderman, Le~Scao, Bari, Shen, Yong, Schoelkopf, Tang, Radev, Aji, Almubarak, Albanie, Alyafeai, Webson, Raff, and Raffel}]{muennighoff2022crosslingual}
Niklas Muennighoff, Thomas Wang, Lintang Sutawika, Adam Roberts, Stella Biderman, Teven Le~Scao, M~Saiful Bari, Sheng Shen, Zheng~Xin Yong, Hailey Schoelkopf, Xiangru Tang, Dragomir Radev, Alham~Fikri Aji, Khalid Almubarak, Samuel Albanie, Zaid Alyafeai, Albert Webson, Edward Raff, and Colin Raffel. 2023{\natexlab{b}}.
\newblock \href {https://doi.org/10.18653/v1/2023.acl-long.891} {Crosslingual generalization through multitask finetuning}.
\newblock In \emph{Proc. of the 61st ACL}, pages 15991--16111, Toronto, Canada.

\bibitem[{Neelakantan et~al.(2022)Neelakantan, Xu, Puri, Radford, Han, Tworek, Yuan, Tezak, Kim, Hallacy et~al.}]{neelakantan2022text}
Arvind Neelakantan, Tao Xu, Raul Puri, Alec Radford, Jesse~Michael Han, Jerry Tworek, Qiming Yuan, Nikolas Tezak, Jong~Wook Kim, Chris Hallacy, et~al. 2022.
\newblock \href {https://arxiv.org/abs/2201.10005} {Text and code embeddings by contrastive pre-training}.
\newblock \emph{arXiv preprint arXiv:2201.10005}.

\bibitem[{Nguyen et~al.(2024)Nguyen, Nguyen, Lai, Man, Ngo, Dernoncourt, Rossi, and Nguyen}]{nguyen2023culturax}
Thuat Nguyen, Chien~Van Nguyen, Viet~Dac Lai, Hieu Man, Nghia~Trung Ngo, Franck Dernoncourt, Ryan~A. Rossi, and Thien~Huu Nguyen. 2024.
\newblock \href {https://aclanthology.org/2024.lrec-main.377} {{C}ultura{X}: A cleaned, enormous, and multilingual dataset for large language models in 167 languages}.
\newblock In \emph{Proc. of the 2024 LREC-COLING}, pages 4226--4237, Torino, Italia. ELRA and ICCL.

\bibitem[{Nguyen et~al.(2016)Nguyen, Rosenberg, Song, Gao, Tiwary, Majumder, and Deng}]{DBLP:conf/nips/NguyenRSGTMD16}
Tri Nguyen, Mir Rosenberg, Xia Song, Jianfeng Gao, Saurabh Tiwary, Rangan Majumder, and Li~Deng. 2016.
\newblock \href {https://ceur-ws.org/Vol-1773/CoCoNIPS\_2016\_paper9.pdf} {{MS} {MARCO:} {A} human generated machine reading comprehension dataset}.
\newblock In \emph{Proc. of the Workshop on Cognitive Computation: Integrating neural and symbolic approaches 2016}, volume 1773 of \emph{{CEUR} Workshop Proceedings}.

\bibitem[{Nussbaum et~al.(2024)Nussbaum, Morris, Duderstadt, and Mulyar}]{nussbaum2024nomic}
Zach Nussbaum, John~X Morris, Brandon Duderstadt, and Andriy Mulyar. 2024.
\newblock \href {https://arxiv.org/abs/2402.01613} {Nomic embed: Training a reproducible long context text embedder}.
\newblock \emph{arXiv preprint arXiv:2402.01613}.

\bibitem[{Oord et~al.(2018)Oord, Li, and Vinyals}]{oord2018representation}
Aaron van~den Oord, Yazhe Li, and Oriol Vinyals. 2018.
\newblock \href {https://arxiv.org/abs/1807.03748} {Representation learning with contrastive predictive coding}.
\newblock \emph{arXiv preprint arXiv:1807.03748}.

\bibitem[{Ponti et~al.(2020)Ponti, Glava{\v{s}}, Majewska, Liu, Vuli{\'c}, and Korhonen}]{ponti-etal-2020-xcopa}
Edoardo~Maria Ponti, Goran Glava{\v{s}}, Olga Majewska, Qianchu Liu, Ivan Vuli{\'c}, and Anna Korhonen. 2020.
\newblock \href {https://doi.org/10.18653/v1/2020.emnlp-main.185} {{XCOPA}: A multilingual dataset for causal commonsense reasoning}.
\newblock In \emph{Proc. of the EMNLP}, pages 2362--2376, Online.

\bibitem[{Portes et~al.(2023)Portes, Trott, Havens, King, Venigalla, Nadeem, Sardana, Khudia, and Frankle}]{portes2023mosaicbert}
Jacob Portes, Alexander~R Trott, Sam Havens, Daniel King, Abhinav Venigalla, Moin Nadeem, Nikhil Sardana, Daya Khudia, and Jonathan Frankle. 2023.
\newblock \href {https://papers.nips.cc/paper_files/paper/2023/file/095a6917768712b7ccc61acbeecad1d8-Paper-Conference.pdf} {{MosaicBERT}: A bidirectional encoder optimized for fast pretraining}.
\newblock In \emph{Thirty-seventh Conference on Neural Information Processing Systems}.

\bibitem[{Po{\'s}wiata et~al.(2024)Po{\'s}wiata, Dadas, and Pere{\l}kiewicz}]{poswiata2024pl}
Rafa{\l} Po{\'s}wiata, S{\l}awomir Dadas, and Micha{\l} Pere{\l}kiewicz. 2024.
\newblock \href {https://arxiv.org/abs/2405.10138} {Pl-mteb: Polish massive text embedding benchmark}.
\newblock \emph{arXiv preprint arXiv:2405.10138}.

\bibitem[{Press et~al.(2022)Press, Smith, and Lewis}]{press2022train}
Ofir Press, Noah Smith, and Mike Lewis. 2022.
\newblock \href {https://openreview.net/forum?id=R8sQPpGCv0} {Train short, test long: Attention with linear biases enables input length extrapolation}.
\newblock In \emph{International Conference on Learning Representations}.

\bibitem[{Qiu et~al.(2022)Qiu, Li, Qu, Chen, She, Liu, Wu, and Wang}]{qiu-etal-2022-dureader}
Yifu Qiu, Hongyu Li, Yingqi Qu, Ying Chen, QiaoQiao She, Jing Liu, Hua Wu, and Haifeng Wang. 2022.
\newblock \href {https://doi.org/10.18653/v1/2022.emnlp-main.357} {{D}u{R}eader-retrieval: A large-scale {C}hinese benchmark for passage retrieval from web search engine}.
\newblock In \emph{Proc. of the EMNLP}, pages 5326--5338, Abu Dhabi, United Arab Emirates.

\bibitem[{Rabe and Staats(2021)}]{rabe2021self}
Markus~N Rabe and Charles Staats. 2021.
\newblock \href {https://arxiv.org/abs/2112.05682} {Self-attention does not need o(n2) memory}.
\newblock \emph{arXiv preprint arXiv:2112.05682}.

\bibitem[{Raffel et~al.(2020)Raffel, Shazeer, Roberts, Lee, Narang, Matena, Zhou, Li, and Liu}]{raffel2020exploring}
Colin Raffel, Noam Shazeer, Adam Roberts, Katherine Lee, Sharan Narang, Michael Matena, Yanqi Zhou, Wei Li, and Peter~J Liu. 2020.
\newblock \href {https://www.jmlr.org/papers/v21/20-074.html} {Exploring the limits of transfer learning with a unified text-to-text transformer}.
\newblock \emph{Journal of machine learning research}, 21(140):1--67.

\bibitem[{Rahimi et~al.(2019)Rahimi, Li, and Cohn}]{rahimi-etal-2019-massively}
Afshin Rahimi, Yuan Li, and Trevor Cohn. 2019.
\newblock \href {https://www.aclweb.org/anthology/P19-1015} {Massively multilingual transfer for {NER}}.
\newblock In \emph{Proc. of the ACL}, pages 151--164, Florence, Italy.

\bibitem[{Rajbhandari et~al.(2020)Rajbhandari, Rasley, Ruwase, and He}]{rajbhandari2020zero}
Samyam Rajbhandari, Jeff Rasley, Olatunji Ruwase, and Yuxiong He. 2020.
\newblock \href {https://ieeexplore.ieee.org/abstract/document/9355301/} {Zero: Memory optimizations toward training trillion parameter models}.
\newblock In \emph{SC20: International Conference for High Performance Computing, Networking, Storage and Analysis}, pages 1--16. IEEE.

\bibitem[{Rajpurkar et~al.(2016)Rajpurkar, Zhang, Lopyrev, and Liang}]{rajpurkar-etal-2016-squad}
Pranav Rajpurkar, Jian Zhang, Konstantin Lopyrev, and Percy Liang. 2016.
\newblock \href {https://doi.org/10.18653/v1/D16-1264} {{SQ}u{AD}: 100,000+ questions for machine comprehension of text}.
\newblock In \emph{Proc. of the EMNLP}, pages 2383--2392, Austin, Texas.

\bibitem[{Roemmele et~al.(2011)Roemmele, Bejan, and Gordon}]{roemmele2011choice}
Melissa Roemmele, Cosmin~Adrian Bejan, and Andrew~S Gordon. 2011.
\newblock \href {https://cdn.aaai.org/ocs/2418/2418-10878-1-PB.pdf} {Choice of plausible alternatives: An evaluation of commonsense causal reasoning}.
\newblock In \emph{2011 AAAI spring symposium series}.

\bibitem[{Roy et~al.(2020)Roy, Constant, Al-Rfou, Barua, Phillips, and Yang}]{roy-etal-2020-lareqa}
Uma Roy, Noah Constant, Rami Al-Rfou, Aditya Barua, Aaron Phillips, and Yinfei Yang. 2020.
\newblock \href {https://doi.org/10.18653/v1/2020.emnlp-main.477} {{LAR}e{QA}: Language-agnostic answer retrieval from a multilingual pool}.
\newblock In \emph{Proc. of the EMNLP 2020}, pages 5919--5930, Online.

\bibitem[{Ruder et~al.(2021)Ruder, Constant, Botha, Siddhant, Firat, Fu, Liu, Hu, Garrette, Neubig, and Johnson}]{ruder-etal-2021-xtreme}
Sebastian Ruder, Noah Constant, Jan Botha, Aditya Siddhant, Orhan Firat, Jinlan Fu, Pengfei Liu, Junjie Hu, Dan Garrette, Graham Neubig, and Melvin Johnson. 2021.
\newblock \href {https://doi.org/10.18653/v1/2021.emnlp-main.802} {{XTREME}-{R}: Towards more challenging and nuanced multilingual evaluation}.
\newblock In \emph{Proc. of the EMNLP 2021}, pages 10215--10245, Online and Punta Cana, Dominican Republic.

\bibitem[{Saad{-}Falcon et~al.(2024)Saad{-}Falcon, Fu, Arora, Guha, and R{\'{e}}}]{saadbenchmarking}
Jon Saad{-}Falcon, Daniel~Y. Fu, Simran Arora, Neel Guha, and Christopher R{\'{e}}. 2024.
\newblock \href {https://openreview.net/forum?id=HkCRgoGtt6} {Benchmarking and building long-context retrieval models with loco and {M2-BERT}}.
\newblock In \emph{Forty-first International Conference on Machine Learning}.

\bibitem[{Shazeer(2020)}]{shazeer2020glu}
Noam Shazeer. 2020.
\newblock \href {https://arxiv.org/abs/2002.05202} {Glu variants improve transformer}.
\newblock \emph{arXiv preprint arXiv:2002.05202}.

\bibitem[{Socher et~al.(2013)Socher, Perelygin, Wu, Chuang, Manning, Ng, and Potts}]{socher-etal-2013-recursive}
Richard Socher, Alex Perelygin, Jean Wu, Jason Chuang, Christopher~D. Manning, Andrew Ng, and Christopher Potts. 2013.
\newblock \href {https://aclanthology.org/D13-1170} {Recursive deep models for semantic compositionality over a sentiment treebank}.
\newblock In \emph{Proc. of the EMNLP}, pages 1631--1642, Seattle, Washington, USA.

\bibitem[{Su et~al.(2024)Su, Ahmed, Lu, Pan, Bo, and Liu}]{su2024roformer}
Jianlin Su, Murtadha Ahmed, Yu~Lu, Shengfeng Pan, Wen Bo, and Yunfeng Liu. 2024.
\newblock \href {https://www.sciencedirect.com/science/article/pii/S0925231223011864} {Roformer: Enhanced transformer with rotary position embedding}.
\newblock \emph{Neurocomputing}, 568:127063.

\bibitem[{Team et~al.(2024)}]{nllb2024scaling}
NLLB Team et~al. 2024.
\newblock \href {https://www.nature.com/articles/s41586-024-07335-x} {Scaling neural machine translation to 200 languages}.
\newblock \emph{Nature}, 630(8018):841.

\bibitem[{Thakur et~al.(2021)Thakur, Reimers, R{\"u}ckl{\'e}, Srivastava, and Gurevych}]{thakur2beir}
Nandan Thakur, Nils Reimers, Andreas R{\"u}ckl{\'e}, Abhishek Srivastava, and Iryna Gurevych. 2021.
\newblock \href {https://neurips.cc/media/neurips-2021/Slides/29903.pdf} {Beir: A heterogeneous benchmark for zero-shot evaluation of information retrieval models}.
\newblock In \emph{Thirty-fifth Conference on Neural Information Processing Systems Datasets and Benchmarks Track (Round 2)}.

\bibitem[{Thorne et~al.(2018)Thorne, Vlachos, Christodoulopoulos, and Mittal}]{thorne-etal-2018-fever}
James Thorne, Andreas Vlachos, Christos Christodoulopoulos, and Arpit Mittal. 2018.
\newblock \href {https://doi.org/10.18653/v1/N18-1074} {{FEVER}: a large-scale dataset for fact extraction and {VER}ification}.
\newblock In \emph{Proc. of the NAACL-HLT}, pages 809--819, New Orleans, Louisiana.

\bibitem[{Touvron et~al.(2023)Touvron, Martin, Stone, Albert, Almahairi, Babaei, Bashlykov, Batra, Bhargava, Bhosale et~al.}]{touvron2023llama}
Hugo Touvron, Louis Martin, Kevin Stone, Peter Albert, Amjad Almahairi, Yasmine Babaei, Nikolay Bashlykov, Soumya Batra, Prajjwal Bhargava, Shruti Bhosale, et~al. 2023.
\newblock \href {https://arxiv.org/abs/2307.09288} {Llama 2: Open foundation and fine-tuned chat models}.
\newblock \emph{arXiv preprint arXiv:2307.09288}.

\bibitem[{Wang et~al.(2018)Wang, Singh, Michael, Hill, Levy, and Bowman}]{wang2018glue}
Alex Wang, Amanpreet Singh, Julian Michael, Felix Hill, Omer Levy, and Samuel~R Bowman. 2018.
\newblock \href {https://openreview.net/forum?id=rJ4km2R5t7} {Glue: A multi-task benchmark and analysis platform for natural language understanding}.
\newblock In \emph{International Conference on Learning Representations}.

\bibitem[{Wang et~al.(2022)Wang, Yang, Huang, Jiao, Yang, Jiang, Majumder, and Wei}]{wang2022text}
Liang Wang, Nan Yang, Xiaolong Huang, Binxing Jiao, Linjun Yang, Daxin Jiang, Rangan Majumder, and Furu Wei. 2022.
\newblock \href {https://arxiv.org/abs/2212.03533} {Text embeddings by weakly-supervised contrastive pre-training}.
\newblock \emph{arXiv preprint arXiv:2212.03533}.

\bibitem[{Wang et~al.(2024{\natexlab{a}})Wang, Yang, Huang, Yang, Majumder, and Wei}]{wang2024improving}
Liang Wang, Nan Yang, Xiaolong Huang, Linjun Yang, Rangan Majumder, and Furu Wei. 2024{\natexlab{a}}.
\newblock \href {https://doi.org/10.18653/v1/2024.acl-long.642} {Improving text embeddings with large language models}.
\newblock In \emph{Proc. of the 62nd ACL}, pages 11897--11916, Bangkok, Thailand. Association for Computational Linguistics.

\bibitem[{Wang et~al.(2024{\natexlab{b}})Wang, Yang, Huang, Yang, Majumder, and Wei}]{wang2024multilingual}
Liang Wang, Nan Yang, Xiaolong Huang, Linjun Yang, Rangan Majumder, and Furu Wei. 2024{\natexlab{b}}.
\newblock \href {https://arxiv.org/abs/2402.05672} {Multilingual e5 text embeddings: A technical report}.
\newblock \emph{arXiv preprint arXiv:2402.05672}.

\bibitem[{Warstadt et~al.(2019)Warstadt, Singh, and Bowman}]{warstadt-etal-2019-neural}
Alex Warstadt, Amanpreet Singh, and Samuel~R. Bowman. 2019.
\newblock \href {https://doi.org/10.1162/tacl_a_00290} {Neural network acceptability judgments}.
\newblock \emph{Transactions of the Association for Computational Linguistics}, 7:625--641.

\bibitem[{Wei et~al.(2023)Wei, Zhao, Zhang, Zhu, Wang, Yang, Li, Cheng, L{\"u}, Hu et~al.}]{wei2023skywork}
Tianwen Wei, Liang Zhao, Lichang Zhang, Bo~Zhu, Lijie Wang, Haihua Yang, Biye Li, Cheng Cheng, Weiwei L{\"u}, Rui Hu, et~al. 2023.
\newblock \href {https://arxiv.org/abs/2310.19341} {Skywork: A more open bilingual foundation model}.
\newblock \emph{arXiv preprint arXiv:2310.19341}.

\bibitem[{Williams et~al.(2018)Williams, Nangia, and Bowman}]{williams-etal-2018-broad}
Adina Williams, Nikita Nangia, and Samuel Bowman. 2018.
\newblock \href {https://doi.org/10.18653/v1/N18-1101} {A broad-coverage challenge corpus for sentence understanding through inference}.
\newblock In \emph{Proc. of the NAACL-HLT}, pages 1112--1122, New Orleans, Louisiana.

\bibitem[{Wolf et~al.(2020)Wolf, Debut, Sanh, Chaumond, Delangue, Moi, Cistac, Rault, Louf, Funtowicz, Davison, Shleifer, von Platen, Ma, Jernite, Plu, Xu, Scao, Gugger, Drame, Lhoest, and Rush}]{wolf-etal-2020-transformers}
Thomas Wolf, Lysandre Debut, Victor Sanh, Julien Chaumond, Clement Delangue, Anthony Moi, Pierric Cistac, Tim Rault, Rémi Louf, Morgan Funtowicz, Joe Davison, Sam Shleifer, Patrick von Platen, Clara Ma, Yacine Jernite, Julien Plu, Canwen Xu, Teven~Le Scao, Sylvain Gugger, Mariama Drame, Quentin Lhoest, and Alexander~M. Rush. 2020.
\newblock \href {https://www.aclweb.org/anthology/2020.emnlp-demos.6} {Transformers: State-of-the-art natural language processing}.
\newblock In \emph{Proc. of the EMNLP 2020: System Demonstrations}, pages 38--45, Online.

\bibitem[{Xiao et~al.(2022)Xiao, Liu, Shao, and Cao}]{xiao-etal-2022-retromae}
Shitao Xiao, Zheng Liu, Yingxia Shao, and Zhao Cao. 2022.
\newblock \href {https://doi.org/10.18653/v1/2022.emnlp-main.35} {{R}etro{MAE}: Pre-training retrieval-oriented language models via masked auto-encoder}.
\newblock In \emph{Proc. of the EMNLP}, pages 538--548, Abu Dhabi, United Arab Emirates.

\bibitem[{Xiao et~al.(2024)Xiao, Liu, Zhang, Muennighoff, Lian, and Nie}]{xiao2023c}
Shitao Xiao, Zheng Liu, Peitian Zhang, Niklas Muennighoff, Defu Lian, and Jian-Yun Nie. 2024.
\newblock \href {https://doi.org/10.1145/3626772.3657878} {C-pack: Packed resources for general chinese embeddings}.
\newblock In \emph{Proc. of the 47th SIGIR}, page 641–649, New York, NY, USA. Association for Computing Machinery.

\bibitem[{Xie et~al.(2023{\natexlab{a}})Xie, Dong, Wang, Lv, Yao, Gan, Wu, Li, Li, Liu, and Ma}]{xie2023t2ranking}
Xiaohui Xie, Qian Dong, Bingning Wang, Feiyang Lv, Ting Yao, Weinan Gan, Zhijing Wu, Xiangsheng Li, Haitao Li, Yiqun Liu, and Jin Ma. 2023{\natexlab{a}}.
\newblock \href {https://doi.org/10.1145/3539618.3591874} {T2ranking: A large-scale chinese benchmark for passage ranking}.
\newblock In \emph{Proceedings of the 46th SIGIR}, page 2681–2690, New York, NY, USA.

\bibitem[{Xie et~al.(2023{\natexlab{b}})Xie, Zhang, Sato, Sugiyama et~al.}]{xie2024overlooked}
Zeke Xie, Jingzhao Zhang, Issei Sato, Masashi Sugiyama, et~al. 2023{\natexlab{b}}.
\newblock \href {https://proceedings.neurips.cc/paper_files/paper/2023/file/040d3b6af368bf71f952c18da5713b48-Paper-Conference.pdf} {On the overlooked pitfalls of weight decay and how to mitigate them: A gradient-norm perspective}.
\newblock In \emph{Thirty-seventh Conference on Neural Information Processing Systems}.

\bibitem[{Xiong et~al.(2024)Xiong, Liu, Molybog, Zhang, Bhargava, Hou, Martin, Rungta, Sankararaman, Oguz, Khabsa, Fang, Mehdad, Narang, Malik, Fan, Bhosale, Edunov, Lewis, Wang, and Ma}]{xiong-etal-2024-effective}
Wenhan Xiong, Jingyu Liu, Igor Molybog, Hejia Zhang, Prajjwal Bhargava, Rui Hou, Louis Martin, Rashi Rungta, Karthik~Abinav Sankararaman, Barlas Oguz, Madian Khabsa, Han Fang, Yashar Mehdad, Sharan Narang, Kshitiz Malik, Angela Fan, Shruti Bhosale, Sergey Edunov, Mike Lewis, Sinong Wang, and Hao Ma. 2024.
\newblock \href {https://aclanthology.org/2024.naacl-long.260} {Effective long-context scaling of foundation models}.
\newblock In \emph{Proc. of the NAACL-HLT}, pages 4643--4663, Mexico City, Mexico.

\bibitem[{Xue et~al.(2021)Xue, Constant, Roberts, Kale, Al-Rfou, Siddhant, Barua, and Raffel}]{xue-etal-2021-mt5}
Linting Xue, Noah Constant, Adam Roberts, Mihir Kale, Rami Al-Rfou, Aditya Siddhant, Aditya Barua, and Colin Raffel. 2021.
\newblock \href {https://doi.org/10.18653/v1/2021.naacl-main.41} {m{T}5: A massively multilingual pre-trained text-to-text transformer}.
\newblock In \emph{Proc. of the NAACL-HLT}, pages 483--498, Online.

\bibitem[{Yang et~al.(2018)Yang, Qi, Zhang, Bengio, Cohen, Salakhutdinov, and Manning}]{yang-etal-2018-hotpotqa}
Zhilin Yang, Peng Qi, Saizheng Zhang, Yoshua Bengio, William Cohen, Ruslan Salakhutdinov, and Christopher~D. Manning. 2018.
\newblock \href {https://doi.org/10.18653/v1/D18-1259} {{H}otpot{QA}: A dataset for diverse, explainable multi-hop question answering}.
\newblock In \emph{Proc. of the EMNLP}, pages 2369--2380, Brussels, Belgium.

\bibitem[{Zhang et~al.(2018)Zhang, Zhang, Wang, Guo, and Liu}]{Zhang2018MultiScaleAI}
Sheng Zhang, Xin Zhang, Hui Wang, Lixiang Guo, and Shanshan Liu. 2018.
\newblock \href {https://api.semanticscholar.org/CorpusID:56598900} {Multi-scale attentive interaction networks for chinese medical question answer selection}.
\newblock \emph{IEEE Access}, 6:74061--74071.

\bibitem[{Zhang et~al.(2023{\natexlab{a}})Zhang, Li, Zhang, Long, Xie, Zhang, and Zhang}]{zhang2023language}
Xin Zhang, Zehan Li, Yanzhao Zhang, Dingkun Long, Pengjun Xie, Meishan Zhang, and Min Zhang. 2023{\natexlab{a}}.
\newblock \href {https://arxiv.org/abs/2310.08232} {Language models are universal embedders}.
\newblock \emph{arXiv preprint arXiv:2310.08232}.

\bibitem[{Zhang et~al.(2021)Zhang, Ma, Shi, and Lin}]{zhang2021mrtydi}
Xinyu Zhang, Xueguang Ma, Peng Shi, and Jimmy Lin. 2021.
\newblock \href {https://doi.org/10.18653/v1/2021.mrl-1.12} {Mr. {T}y{D}i: A multi-lingual benchmark for dense retrieval}.
\newblock In \emph{Proceedings of the 1st Workshop on Multilingual Representation Learning}, pages 127--137, Punta Cana, Dominican Republic.

\bibitem[{Zhang et~al.(2023{\natexlab{b}})Zhang, Thakur, Ogundepo, Kamalloo, Alfonso-Hermelo, Li, Liu, Rezagholizadeh, and Lin}]{10.1162/tacl_a_00595}
Xinyu Zhang, Nandan Thakur, Odunayo Ogundepo, Ehsan Kamalloo, David Alfonso-Hermelo, Xiaoguang Li, Qun Liu, Mehdi Rezagholizadeh, and Jimmy Lin. 2023{\natexlab{b}}.
\newblock \href {https://doi.org/10.1162/tacl_a_00595} {{MIRACL: A Multilingual Retrieval Dataset Covering 18 Diverse Languages}}.
\newblock \emph{Transactions of the Association for Computational Linguistics}, 11:1114--1131.

\bibitem[{Zhang et~al.(2022)Zhang, Long, Xu, and Xie}]{zhang2022hlatr}
Yanzhao Zhang, Dingkun Long, Guangwei Xu, and Pengjun Xie. 2022.
\newblock \href {https://arxiv.org/abs/2205.10569} {Hlatr: enhance multi-stage text retrieval with hybrid list aware transformer reranking}.
\newblock \emph{arXiv preprint arXiv:2205.10569}.

\bibitem[{Zhao et~al.(2024)Zhao, Liu, Ren, and Wen}]{zhao2024dense}
Wayne~Xin Zhao, Jing Liu, Ruiyang Ren, and Ji-Rong Wen. 2024.
\newblock \href {https://doi.org/10.1145/3637870} {Dense text retrieval based on pretrained language models: A survey}.
\newblock \emph{ACM Trans. Inf. Syst.}, 42(4).

\bibitem[{Zhu et~al.(2024)Zhu, Wang, Yang, Song, Wu, Wei, and Li}]{zhu2024longembed}
Dawei Zhu, Liang Wang, Nan Yang, Yifan Song, Wenhao Wu, Furu Wei, and Sujian Li. 2024.
\newblock \href {https://arxiv.org/abs/2404.12096} {Longembed: Extending embedding models for long context retrieval}.
\newblock \emph{arXiv preprint arXiv:2404.12096}.

\end{thebibliography}

\appendix
\section*{Appendix}
\label{sec:appendix}

\section{MLM Pre-Training}
In this section, we describe the data and training configurations of the MLM pre-training of our suggested text encoder.

\subsection{Data}
Our multilingual pre-training data are composed from following sources:
\begin{compactitem}
\item C4 \citep{raffel2020exploring},
\item Skypile \citep{wei2023skywork} (2021-2023 subsets),
\item mC4 \citep{xue-etal-2021-mt5} (excluded English),
\item CulturaX \citep{nguyen2023culturax},
\item Wikipedia \cite{wikidump},
\item books (proprietary).
\end{compactitem}
We filter them and curate a dataset with 1,028B tokens (by XLM-R tokenizer), covering 75 languages (Chinese Simplified and Traditional are counted as one).
Table \ref{tab:mlm-data} presents the statistics of our final dataset.

\begin{table*}
\centering
\resizebox{\textwidth}{!}{
\begin{tabular}[b]{clrrclrr}
\toprule
\textbf{ISO code} & \textbf{Language} & \textbf{Tokens} (M) & \textbf{Size} (GiB) & \textbf{ISO code} & \textbf{Language} & \textbf{Tokens} (M) & \textbf{Size} (GiB)\\
\cmidrule(r){1-4}\cmidrule(l){5-8}
af  & Afrikaans   & 1,489.19   & 5.30   &    ky    & Kyrgyz     & 500.40    & 3.27   \\
ar  & Arabic      & 14,549.36  & 79.53  &    lo    & Lao        & 2.43      & 0.01   \\
az  & Azerbaijani & 688.72     & 3.13   &    lt    & Lithuanian & 1,824.46  & 6.38   \\
be  & Belarusian  & 1,090.61   & 6.17   &    lv    & Latvian    & 1,823.43  & 6.38   \\
bg  & Bulgarian   & 1,454.57   & 8.94   &    mk    & Macedonian & 735.46    & 4.89   \\
bn  & Bengali     & 1,291.58   & 9.21   &    ml    & Malayalam  & 778.66    & 7.27   \\
ca  & Catalan     & 1,294.05   & 4.65   &    mn    & Mongolian  & 958.83    & 5.91   \\
ceb & Cebuano     & 633.06     & 2.02   &    mr    & Marathi    & 861.05    & 7.48   \\
cs  & Czech       & 1,465.00   & 5.27   &    ms    & Malay      & 96.37     & 0.39   \\
cy  & Welsh       & 582.49     & 1.84   &    my    & Burmese    & 902.46    & 7.26   \\
da  & Danish      & 1,030.30   & 4.01   &    ne    & Nepali     & 657.65    & 6.32   \\
de  & German      & 18,097.31  & 67.90  &    nl    & Dutch      & 5,137.98  & 18.65   \\
el  & Greek       & 874.87     & 5.09   &    no    & Norwegian  & 992.51    & 3.91   \\
en  & English     & 187,110.31 & 771.79 &    pa    & Punjabi    & 726.41    & 4.96   \\
es  & Spanish     & 148,713.06 & 601.04 &    pl    & Polish     & 2,949.88  & 10.42   \\
et  & Estonian    & 1,111.31   & 4.10   &    pt    & Portuguese & 49,594.59 & 198.64   \\
eu  & Basque      & 787.46     & 2.99   &    qu    & Quechua    & 0.07      & 0.00   \\
fa  & Persian     & 1,203.16   & 7.22   &    ro    & Romanian   & 2,215.05  & 7.98   \\
fi  & Finnish     & 949.88     & 3.73   &    ru    & Russian    & 93,966.28 & 597.92   \\
fr  & French      & 136,785.00 & 512.28 &    si    & Sinhala    & 878.65    & 7.03   \\
gl  & Galician    & 772.47     & 3.22   &    sk    & Slovak     & 884.38    & 3.31   \\
gu  & Gujarati    & 973.27     & 6.95   &    sl    & Slovenian  & 1,100.81  & 4.05   \\
he  & Hebrew      & 1,842.74   & 8.36   &    so    & Somali     & 0.82      & 0.00   \\
hi  & Hindi       & 1,032.67   & 8.27   &    sq    & Albanian   & 700.78    & 2.73   \\
hr  & Croatian    & 480.19     & 1.54   &    sr    & Serbian    & 1,139.38  & 6.84   \\
ht  & Haitian     & 0.03       & 0.00   &    sv    & Swedish    & 840.00    & 3.37   \\
hu  & Hungarian   & 1,341.23   & 5.10   &    sw    & Swahili    & 31.58     & 0.13   \\
hy  & Armenian    & 805.98     & 4.88   &    ta    & Tamil      & 926.84    & 8.54   \\
id  & Indonesian  & 25,564.33  & 119.84 &    te    & Telugu     & 857.91    & 7.01   \\
is  & Icelandic   & 987.89     & 3.63   &    th    & Thai       & 12,782.08 & 119.52   \\
it  & Italian     & 11,068.23  & 40.50  &    tl    & Filipino   & 275.16    & 1.01   \\
ja  & Japanese    & 135,684.28 & 601.19 &    tr    & Turkish    & 1,065.05  & 4.42   \\
jv  & Javanese    & 0.62       & 0.00   &    uk    & Ukrainian  & 893.70    & 5.68   \\
ka  & Georgian    & 834.90     & 7.25   &    ur    & Urdu       & 1,051.83  & 6.19   \\
kk  & Kazakh      & 1,020.27   & 6.57   &    vi    & Vietnamese & 67,850.87 & 305.51   \\
km  & Khmer       & 746.15     & 6.54   &    yo    & Yoruba     & 0.04      & 0.00   \\
kn  & Kannada     & 919.83     & 7.15   &    zh-cn & Chinese (Simplified)  & 43,727.30 & 167.23 \\
ko  & Korean      & 22,865.85  & 91.78  &    zh-tw & Chinese (Traditional) & 73.39 & 0.26  \\
\bottomrule
\end{tabular}
}
\caption{
MLM pre-training data, where we have a total of 1,028B tokens (by XLM-RoBERTa tokenizer). The raw texts are stored in 4.47 TiB arrow files. We report the list of 75 languages (Chinese Simplified and Traditional are counted as one) and include the number of tokens and the size of the data (arrow files, in GiB) for each language.
}
\label{tab:mlm-data}
\end{table*}

\subsection{Training Details}\label{sec:app:mlm-training}
We pre-train out text encoder with a two-stage curriculum by masked language model (MLM) objective.
The first stage model is trained on maximum length $2048$ with batch size $8192$ for roughly $0.6$ epoch (250k steps) on sampled data (by XLM sampling Eq.\ref{eq:xlm-sampling}).
In the second stage, we down sample texts shorter than $2048$ and continue train the model for 30k steps with maximum length $8192$ and batch size $2048$.
The RoPE base is set to $10,000$ and $160,000$ for the first and second stage, respectively \cite{xiong-etal-2024-effective,liuscaling,men2024base}.

The text encoder is initialized in base size ($12$ layers of hidden state size $768$) by PyTorch default initialization.
We train the model by \texttt{transformers} library \cite{wolf-etal-2020-transformers} in BF16 precision.
Following \citet{portes2023mosaicbert}, we use the learning rate decoupled AdamW optimizer with weight decay 1e-5. 
The other hyper-parameters are in Table \ref{tab:hparam-mlm}.
During training, we split texts that exceed the max sequence length into chunks, but we do not modify shorter texts.

The 250k steps of first stage, MLM-2048, took 10.75 days on 32 A100 80G GPUs.
The 30k steps of second stage, MLM-8192, took 20.5 hours on 32 A100 80G GPUs.
We acknowledge that this is not the optimal setting and recommend further explorations to optimize the pre-training.

\subsection{Additional Discussion on RoPE}
We chose RoPE \cite{su2024roformer} (to replace absolute position embedding) due to its advantageous properties.
RoPE offers excellent context extension capabilities, allowing models to be trained on shorter context windows and then run inference on longer ones.
Additionally, it implements asymmetric relative distance encoding, meaning $D(i, j) \ne D(j, i)$, which appears to be particularly important for the training of BERT-like encoder-only models that rely on bidirectional attention.
Furthermore, the effectiveness of RoPE has been empirically validated by numerous models, such as RoFormer \cite{su2024roformer} and LLaMA \cite{touvron2023llama}.

\begin{table}
\centering
\resizebox{\columnwidth}{!}{
\begin{tabular}[b]{lcc}
\toprule
Hyper-param & MLM-2048 & MLM-8192  \\
\midrule 
Number of Params & \multicolumn{2}{c}{304M} \\
Number of Layers & \multicolumn{2}{c}{12} \\
Hidden Size & \multicolumn{2}{c}{768} \\
FFN Inner Size & \multicolumn{2}{c}{3072} \\
Number of Attention Heads & \multicolumn{2}{c}{12} \\
Attention Head Size & \multicolumn{2}{c}{64}  \\
Dropout & \multicolumn{2}{c}{0.1}  \\
Attention Dropout & \multicolumn{2}{c}{0}  \\
Learning Rate Decay & \multicolumn{2}{c}{Linear}  \\
Adam $\epsilon$ & \multicolumn{2}{c}{1e-6}  \\
Adam $\beta_1$  & \multicolumn{2}{c}{0.9}  \\
Adam $\beta_2$  & \multicolumn{2}{c}{0.98} \\
Gradient Clipping & \multicolumn{2}{c}{0.0} \\
Precision &  \multicolumn{2}{c}{PyTorch BF16 AMP}   \\
Weight Decay & \multicolumn{2}{c}{1e-5}  \\
Max Length & 2048 & 8192  \\
Batch Size & 8192 & 2048  \\
Peak Learning Rate & 5e-4 & 5e-5 \\
Warm-up Ratio & 0.06 & 0.06 \\
Max Steps & 250000 & 30000 \\
RoPE base & 10000 & 160000 \\
\bottomrule
\end{tabular}
}
\caption{
MLM pre-training hyper-parameters.
}
\label{tab:hparam-mlm}
\end{table}

\section{Contrastive Learning}
In this section, we describe the data and training configurations of the contrastive learning of our TRM and reranker.

\subsection{Pre-Training Data}\label{sec:app:cpt-data}
Following previous studies, we create large-scale weakly correlated text pairs from diverse sources.
The data are primarily consisted of four parts: English pairs \cite{wang2022text,li2023towards}, Chinese pairs \cite{li2023towards,xiao2023c}, multilingual pairs (cc-news\footnote{
\url{commoncrawl.org/blog/news-dataset-available}
}), and crosslingual instruction and translation pairs \cite{muennighoff2022crosslingual,nllb2024scaling}.
We filter the data by removing duplicates and low-quality pairs, resulting in a total of 2,938.8M pairs.
Table \ref{tab:unsupervised-data} lists the statistics of our contrastive pre-training data (cc-news is separately presented by languages in Table \ref{tab:cc-news}).

\begin{table*}
\centering
\resizebox{\textwidth}{!}{
\begin{tabular}[b]{llrrllrr}
\toprule
\textbf{Source} & \textbf{Language} & \textbf{Pairs} (M) & \textbf{Size} (GiB) & \textbf{Source} & \textbf{Language} & \textbf{Pairs} (M) & \textbf{Size} (GiB) \\
\cmidrule(r){1-4}\cmidrule(l){5-8}
agnews & English & 1.15  & 0.30                        & stackoverflow\_title\_body & English & 18.01  & 20.49 \\
amazon\_qa & English & 1.10  & 0.37                    & wikihow & English & 0.13  & 0.03 \\
amazon\_review\_title\_body & English & 87.86  & 43.58 & wikipedia & English & 33.17  & 19.39 \\
arxiv\_title\_abstract & English & 2.26  & 2.26        & yahoo\_body\_answer & English & 0.68  & 0.44 \\
baai\_mtp\_en & English & 196.60  & 178.70             & yahoo\_qa & English & 1.20  & 0.55 \\
beir\_dbpedia & English & 4.64  & 1.59                 & yahoo\_question\_body & English & 0.66  & 0.20 \\
beir\_debate & English & 0.38  & 0.63                  & baai\_mtp\_zh & Chinese & 100.13 & 231.42 \\
beir\_pubmed\_title\_abstract & English & 0.13  & 0.19 & baidu\_baike & Chinese & 34.21 & 39.05 \\
biorxiv\_title\_abstract & English & 0.20  & 0.32      & baike\_qa\_train & Chinese & 1.43 & 1.34 \\
clueweb & English & 3.94  & 6.62                       & commoncrawl\_zh & Chinese & 28.42 & 92.79 \\
clueweb\_anchor & English & 4.51  & 7.69               & gpt3\_qa\_all & Chinese & 4.97 & 2.39 \\
cnn\_dailymail& English  & 0.31  & 1.28                & gpt3\_summarization & Chinese & 4.48 & 1.62 \\
commoncrawl & English & 139.94  & 506.84               & medical\_quac\_wenda\_10m & Chinese & 10.00 & 4.55 \\
dpr\_reddit & English & 199.82  & 125.71               & medical\_scholar & Chinese & 8.43 & 7.81 \\
gooaq\_qa & English & 3.01  & 0.97                     & qcl & Chinese & 7.40 & 43.23 \\
hlp\_wikipedia & English & 19.48  & 13.55              & web\_text\_zh\_train & Chinese & 4.12 & 2.07 \\
medrxiv\_title\_abstract & English & 0.20  & 0.32      & wikipedia & Chinese & 4.45 & 1.07 \\
msmarco & English & 2.89  & 19.56                      & wodao & Chinese & 59.13 & 190.29 \\
npr & English & 0.59  & 1.03                           & zh\_sft\_data\_v1 & Chinese & 0.45 & 0.43 \\
reddit\_title\_body & English & 124.89  & 90.36        & zh\_sft\_data\_v2 & Chinese & 2.24 & 1.37 \\
s2orc\_citation\_abstract & English & 30.58  & 67.81   & zhihu\_qa & Chinese & 53.42 & 40.99 \\
s2orc\_citation\_title & English & 51.03  & 10.84      & zhihu\_title\_body & Chinese & 0.94 & 0.29 \\
s2orc\_title\_abstract & English & 41.77  & 30.29      & xp3x & Crosslingual & 351.87  & 463.85 \\
stackexchange\_qa & English & 3.00  & 3.36             & translation\_eg\_NLLB & Crosslingual & 940.63 & 323.06 \\
stackexchange\_title\_body & English & 4.74  & 4.00 \\
\bottomrule
\end{tabular}
}
\caption{
Contrastive pre-training data, where cc-news multilingual data are not included (Table \ref{tab:cc-news}).
For this Table, we have a total of 2,595.57M pairs (raw texts stored by 2.55 TiB jsonl files).
}
\label{tab:unsupervised-data}
\end{table*}

\begin{table*}
\centering
\resizebox{\textwidth}{!}{
\begin{tabular}[b]{crrcrrcrrcrr}
\toprule
\textbf{Lang.} & \textbf{Pairs} (M) & \textbf{Size} (GiB) &
\textbf{Lang.} & \textbf{Pairs} (M) & \textbf{Size} (GiB) &
\textbf{Lang.} & \textbf{Pairs} (M) & \textbf{Size} (GiB) &
\textbf{Lang.} & \textbf{Pairs} (M) & \textbf{Size} (GiB) \\
ar & 20.407 & 32.45  & fy & 0.044  & 0.03    & lb & 0.048  & 0.05    & sk & 1.093  & 1.16  \\
az & 0.401  & 0.23   & gl & 0.114  & 0.20    & lt & 0.321  & 0.24    & sl & 1.046  & 0.93  \\
be & 0.039  & 0.06   & gu & 0.061  & 0.06    & lv & 0.438  & 0.37    & sq & 0.282  & 0.51  \\
bg & 3.005  & 5.03   & he & 0.397  & 0.84    & mk & 0.173  & 0.44    & sr & 0.910  & 1.09  \\
bn & 0.463  & 0.33   & hi & 14.253 & 29.90   & ml & 0.408  & 0.48    & sv & 3.361  & 2.90  \\
ca & 0.909  & 1.30   & hr & 1.268  & 1.77    & mr & 0.278  & 0.35    & sw & 0.059  & 0.07  \\
cs & 1.834  & 2.18   & hu & 2.668  & 3.40    & my & 0.045  & 0.04    & ta & 2.125  & 1.26  \\ 
da & 1.090  & 1.58   & hy & 0.125  & 0.09    & nl & 6.700  & 7.41    & te & 0.355  & 0.33  \\ 
de & 39.715 & 57.98  & id & 6.048  & 7.46    & nn & 0.162  & 0.12    & tg & 0.038  & 0.03  \\ 
el & 7.170  & 14.93  & is & 0.100  & 0.05    & no & 1.978  & 2.21    & th & 0.124  & 0.17  \\
en & 0.615  & 1.47   & it & 27.827 & 40.57   & or & 0.038  & 0.03    & tl & 0.055  & 0.07  \\
es & 55.201 & 86.87  & ja & 4.139  & 3.95    & pa & 0.036  & 0.04    & tr & 23.840 & 26.81  \\ 
et & 0.950  & 0.85   & ka & 0.074  & 0.06    & pl & 3.530  & 5.77    & uk & 5.021  & 8.42  \\
eu & 0.051  & 0.02   & kn & 0.192  & 0.16    & pt & 12.611 & 19.28   & ur & 1.625  & 0.87  \\
fa & 4.839  & 7.99   & ko & 8.605  & 12.48   & ro & 6.678  & 9.15    & vi & 4.375  & 7.03 \\
fi & 1.532  & 1.93   & ky & 0.061  & 0.03    & ru & 39.451 & 65.74   & MIX$^*$ & 0.359 & 0.28 \\
fr & 21.242 & 32.67  & la & 0.035  & 0.06    & sh & 0.220  & 0.18    & \\ 
\bottomrule
\end{tabular}
}
\caption{
The cc-news multilingual pairs (343.26M in total, raw texts stored by 512.8 GiB jsonl files), used in contrastive pre-training together with all data of Table \ref{tab:unsupervised-data}.
MIX$^*$ denotes the mixed pairs of languages that are less than 1GiB (such as af, ceb).
We utilize a very large batch size ($16,384$), and since each batch contains text exclusively from a single source, these low-resource languages might not fill an entire batch.
Consequently, we have merged these languages together.
}
\label{tab:cc-news}
\end{table*}

\subsection{Fine-Tuning Data}\label{sec:app:cft-data}
We collect publicly available high-quality dataset as our fine-tune data as detailed in Table~\ref{tab:ft-data}. For English, we utilize seven datasets: MS MARCO \cite{DBLP:conf/nips/NguyenRSGTMD16}, Natural Questions (NQ) \cite{kwiatkowski-etal-2019-natural}, TriviaQA \cite{joshi-etal-2017-triviaqa}, HotpotQA \cite{yang-etal-2018-hotpotqa}, SQuAD \cite{rajpurkar-etal-2016-squad}, FEVER \cite{thorne-etal-2018-fever}, AllNLI from SimCSE \cite{gao-etal-2021-simcse}.
For Chinese, we compile six datasets: DuReader~\cite{qiu-etal-2022-dureader}, mMARCO-zh~\cite{bonifacio2021mmarco}, T2-Ranking~\cite{xie2023t2ranking}, CmedQAv2~\cite{Zhang2018MultiScaleAI}, SimCLUE\footnote{\url{https://github.com/CLUEbenchmark/SimCLUE}}, Multi-CPR~\cite{long2022multi}.
Additionally, we incorporate three multilingual datasets: Mr.TyDi~\cite{zhang2021mrtydi}, MIRACL~\cite{10.1162/tacl_a_00595}, and MLDR~\cite{chen-etal-2024-m3}. We exclusively use the trainset of each dataset and employ our contrastive pre-trained model to mine hard negatives.

\begin{table}
\centering
\resizebox{\columnwidth}{!}{
\begin{tabular}[b]{l|c|c}
\toprule
Dataset & Language & Size  \\
\midrule 
MS MARCO, HotpotQA, NQ, NLI, etc. & English & 1.4M  \\
\midrule
DuReader, T$^2$-Ranking, SimCLUE, etc. & Chinese & 2.0M \\
\midrule
MIRACL, Mr.TyDi, MLDR & Multilingual & 118.9K \\
\bottomrule
\end{tabular}
}
\caption{
Specification of training data adopted in Fine-tuning stage.
}
\label{tab:ft-data}
\end{table}

\subsection{TRM Training Setup}\label{sec:app:trm-training}
Here we separately describe the training setting of the contrastive pre-training and TRM fine-tuning.

\paragraph{Contrastive Pre-Training}
In the contrastive pre-training, we train a dense representation model (embedder) which take the [CLS] hidden state as the embedding of the input.
We use the same XLM sampling strategy (eq.\ref{eq:xlm-sampling}) to sample batches from each source of Table \ref{tab:unsupervised-data} or cc-news subset of Table \ref{tab:cc-news}, where the texts of one batch only come from one single source, and the batch size is $16,384$.
We train the model by \texttt{transformers} with \texttt{deepspeed} ZeRO \cite{rajbhandari2020zero} stage 1 in FP16 precision for roughly 0.4 epoch (240k steps, took 154 hours on 16 A100 80G GPUs) of our data (3.93B pairs on sampled data by Eq.\ref{eq:xlm-sampling}).
We use the AdamW optimizer with the learning rate 2e-4, linear decay, and warm-up ratio 0.05.
The $\beta_1=0.9$, $\beta_2=0.999$, and $\epsilon=1e-07$.
We set gradient clipping to 1.0.

\paragraph{TRM Fine-Tuning}

In the fine-tuning stage, we further train our embedding model with high-quality datasets as detailed in \S\ref{sec:app:cft-data}. For each query, we incorporate one positive passage and 8 hard negative passages. To enhance long-context retrieval capabilities and maximize training efficiency, we adopt a dynamic batch size strategy as previous work~\cite{chen-etal-2024-m3}. Firstly, we group the training data according to their lengths for each dataset. Different batch sizes are then used for varying lengths during training. Additionally, we divide the entire batch into multiple sub-batches, encoding each sub-batch iteratively with gradient checkpointing~\cite{chen2016training} and then gather them to get the final batch's embeddings. We train the embedding model with 10 epochs with 8 A100 80G GPUs. All other hyper-parameters remain consistent with those used in the contrastive pre-training stage. In Table~\ref{tab:batch_size}, we list the batch size of different length.

\subsection{Reranker Training Setup}\label{sec:app:reranker-training}

We utilize the identical fine-tuning dataset for both the reranker and the TRM. For each query, we introduce 10 negative samples, comprising 6 hard negatives and 4 randomly selected negatives. All training parameters expect batch size are kept consistent with those employed for the TRM. The batch sizes are listed in Table~\ref{tab:batch_size}.

\begin{table}[]
\resizebox{\columnwidth}{!}{
\begin{tabular}{@{}c|cccc@{}}
\toprule
length & BS(E) & S-BS(R) & BS(E) & S-BS(R) \\ \midrule
0-500     & 768 & 256 & 512 & 256 \\ \midrule
500-1000  & 384 & 128 & 384 & 128 \\ \midrule
1000-2000 & 256 & 64  & 256 & 64  \\ \midrule
2000-3000 & 160 & 48  & 160 & 48  \\ \midrule
3000-8000 & 80  & 16  & 80  & 16  \\ \bottomrule
\end{tabular}
}
\caption{Batch size (BS) and sub batch size (S-BS) of different length for embedding (E) and reranker (R) model in the fine-tune stage.}
\label{tab:batch_size}
\end{table}

\section{NLU Evaluation}
We evaluate our text encoder as well as baselines on the multilingual XTREME-R \cite{ruder-etal-2021-xtreme} and English GLUE \cite{wang2018glue} benchmarks.
We describe the fine-tuning setup and the evaluation details in the following subsections.
The evaluation scripts are available in our github repo\footnote{
\url{github.com/izhx/nlu-evals}
}.

\subsection{XTREME-R}\label{sec:app:xtremer}
We only run XTREME-R \cite{ruder-etal-2021-xtreme} in the zero-shot cross-lingual transfer learning setting, where models are fine-tuned on English trainset and tested on multi- and cross-lingual data.
We compare our encoder with mBERT-base-cased\footnote{
\url{hf.co/google-bert/bert-base-multilingual-cased}
} and XLM-RoBERTa-base\footnote{
\url{hf.co/FacebookAI/xlm-roberta-base}
}.
All models are fine-tuned in the same setting and hyper-parameters.

The results are already presented in Table \ref{tab:xtremer}.

As XTREME-R has no final release, we implement the evaluation code based on the code of XTREME\footnote{\url{github.com/google-research/xtreme}}.
However, there are some differences in the retrieval evaluation, where our code will deduplicate the retrieval corpus.
In addition, we implement the XCOPA in multiple choice, which might be different from XTREME-R.
In fine-tuning, if not specified, we use the epoch number of 3, learning rate of 2e-5, batch size of 32, and max sequence length of 128 \cite{hu2020xtreme}.

\paragraph{XNLI}
We fine-tune the model on MNLI\footnote{
\url{hf.co/datasets/nyu-mll/glue} MNLI subset.
} \cite{williams-etal-2018-broad} trainset and then evaluate the checkpoint on XNLI\footnote{
\url{hf.co/datasets/facebook/xnli}
} \cite{conneau-etal-2018-xnli}.

\paragraph{XCOPA}
We run this data as the multiple choice task.
The model is first trained on SIQA\footnote{
\url{hf.co/datasets/allenai/social\_i\_qa}
} cite{sap-etal-2019-social} and then COPA\footnote{
\url{hf.co/datasets/aps/super\_glue} copa split.
} \cite{roemmele2011choice} for 5 epochs on each dataset.
The checkpoint of COPA is evaluated on XCOPA\footnote{
\url{hf.co/datasets/cambridgeltl/xcopa}
} \cite{ponti-etal-2020-xcopa}.

\paragraph{UDPOS}
We extract pos-tagging data from the UD \cite{de2021universal} v2.7 and train the model on trainset of English parts by 10 epochs.

\paragraph{WikiANN}
We fine-tune the model on the trainset of English by 10 epochs and evaluate on selected WikiANN \cite{rahimi-etal-2019-massively} testsets\footnote{
\url{hf.co/datasets/unimelb-nlp/wikiann}
}.

\paragraph{XQuAD}
We fine-tune on the trainset of SQuAD \cite{rajpurkar-etal-2016-squad} v1.1\footnote{
\url{hf.co/datasets/rajpurkar/squad}
} for 3 epochs with the learning rate 3e-5 and max length 384.
Then we evaluate the checkpoint on XQuAD\footnote{
\url{hf.co/datasets/google/xquad}
} \cite{artetxe-etal-2020-cross}.

\paragraph{MLQA}
We directly evaluate the same checkpoint of XQuAD on MLQA\footnote{
\url{hf.co/datasets/facebook/mlqa}
} \cite{lewis-etal-2020-mlqa} with the same setting.

\paragraph{TyDiQA-GoldP}
We train the model on TyDiQA-GoldP\footnote{
\url{hf.co/datasets/juletxara/tydiqa_xtreme}
} \cite{clark-etal-2020-tydi} trainset in the same setting as XQuAD.
Then we evaluate the checkpoint on the testset.

\paragraph{Mewsli-X}
We generate the data following their github\footnote{
\url{https://github.com/google-research/google-research/blob/master/dense\_representations\_for\_entity\_retrieval/mel/mewsli-x.md\#getting-started}
}.
This is a updated version so that we can not compare with the results in the XTREME-R paper.
We train the model on the English wikipedia (mention, entity)-pairs for 2 epochs with the batch size 64 and max length 64.
Then we evaluate the checkpoint in the language agnostic retrieval setting, refer to \citet{ruder-etal-2021-xtreme} for more details.

\paragraph{LAReQA}
This task is actually conducted on XQuAD-R\footnote{
\url{hf.co/datasets/google-research-datasets/xquad_r}
} \cite{roy-etal-2020-lareqa}.
We fine-tune the model on the trainset of SQuAD v1.1 in dual-encoder architecture ([CLS] as the embedding) and retrieval setting for 3 epochs with the batch size 16, max query length 96, and max document length 256.
Then we evaluate the checkpoint on XQuAD-R in same setting.

\paragraph{Tatoeba}
We directly evaluate the checkpoint from LAReQA on Tatoeba\footnote{
\url{hf.co/datasets/mteb/tatoeba-bitext-mining}
} \cite{tatoebav1} in the same setting.

\subsection{GLUE}\label{sec:app:glue}
The GLUE benchmark \cite{wang2018glue} is English transfer learning, \ie models are trained and tested on the trainset and testset of each dataset (CoLA \cite{warstadt-etal-2019-neural}, SST-2 \cite{socher-etal-2013-recursive}, MRPC \cite{dolan2005automatically}, STS-B \cite{cer-etal-2017-semeval}, QQP, MNLI \cite{williams-etal-2018-broad}, QNLI \cite{rajpurkar-etal-2016-squad}, RTE).

We evaluate the GLUE benchmark based on the scripts\footnote{
\url{github.com/huggingface/transformers/tree/main/examples/pytorch/text-classification\#glue-tasks}
} and data\footnote{
\url{hf.co/datasets/nyu-mll/glue}
} provided by \texttt{transformers}.
In fine-tuning of each dataset, we use the epoch number of 3, learning rate of 2e-5, batch size of 32, and max sequence length of 128.
For MRPC, STS-B, and RTE, we start from the checkpoint of MNLI following \cite{liu2019roberta}.
The MNLI checkpoint is shared with XNLI of XTREME-R (\S\ref{sec:app:xtremer}).

The detailed results are in Table \ref{tab:glue-full}.
We also include scores of our English models (\texttt{Our-en-*}, pre-trained on C4-en) and baselines \cite{portes2023mosaicbert,gunther2023jina2,nussbaum2024nomic}.

\begin{table*}
\centering
\resizebox{\textwidth}{!}{
\begin{tabular}{lcccc|cccccccc}
\toprule
&&&&& \multicolumn{2}{c}{\bf Single Sentence} & \multicolumn{3}{c}{\bf Paraphrase and Similarity} & \multicolumn{3}{c}{\bf Natural Language Inference} \\
\textbf{Model} & \textbf{Params} & \bf Pos. & \bf Seq. & \bf Avg. & CoLA & SST-2 &  MRPC & STS-B & QQP & MNLI & QNLI & RTE \\
\midrule
RoBERTa-base$^\alpha$         & 125M & Abs.  & 512  & \bf 86.4 & 63.6 & 94.8 & 90.2 & 91.2 & 91.9 & 87.6 & 92.8 & 78.7 \\
MosaicBERT-base-128$^\beta$   & 137M & Alibi & 128  & 85.4 & 58.2 & 93.5 & 89.0 & 90.3 & 92.0 & 85.6 & 91.4 & 83.0 \\
MosaicBERT-base-2048$^\gamma$ & 137M & Alibi & 2048 & 85   & 54 & 93 & 87 & 90 & 92 & 86 & 92 & 82 \\
JinaBERT-base$^\delta$        & 137M & Alibi & 512  & 82.6 & 51.4 & 94.5 & 88.4 & 89.5 & 80.7 & 85.7 & 92.2 & 78.7 \\
nomic-bert-2048$^\gamma$      & 137M & RoPE  & 2048 & 84  & 50 & 93 & 88 & 90 & 92 & 86 & 92 & 82  \\
\bf GTEv1.5-en-base-2048     & 137M & RoPE  & 2048 & 85.15 & 54.46 & 93.81 & 93.21 & 90.00 & 88.61 & 86.73 & 91.67 & 82.67 \\
\bf GTEv1.5-en-base-8192     & 137M & RoPE  & 8192 & 85.61 & 57.02 & 93.35 & 92.14 & 90.21 & 88.78 & 86.69 & 91.85 & 84.84 \\
\midrule
XLM-R-base           & 279M & Abs.  & 512  & 80.44 & 30.74 & 92.43 & 92.74 & 89.16 & 87.74 & 84.54 & 90.37 & 75.81 \\
\bf \modelname-MLM-2048    & 305M & RoPE  & 2048 & 83.42 & 49.65 & 92.66 & 91.17 & 89.95 & 88.41 & 85.40 & 91.38 & 78.70 \\
\bf \modelname-MLM-8192    & 305M & RoPE  & 8192 & \bf 83.47 & 48.41 & 92.32 & 90.94 & 89.77 & 88.50 & 85.58 & 91.34 & 80.87 \\
\midrule
RoBERTa-large$^\alpha$       & 355M & Abs.  & 512  & \bf 88.9 & 68.0 & 96.4 & 90.9 & 92.4 & 92.2 & 90.2 & 94.7 & 86.6 \\
MosaicBERT-large-128$^\beta$ & 434M & Alibi & 128  & 86.1 & 59.7 & 93.7 & 88.2 & 90.9 & 92.0 & 86.9 & 93.0 & 84.5 \\
JinaBERT-large$^\delta$      & 435M & Alibi & 512  & 83.7 & 59.6 & 95.0 & 88.5 & 88.2 & 80.9 & 86.6 & 92.5 & 78.5 \\
\bf GTEv1.5-en-large-512        & 434M & RoPE  & 512  & 88.16 & 64.80 & 94.50 & 92.09 & 91.50 & 89.23 & 89.12 & 93.78 & 90.25 \\
\bf GTEv1.5-en-large-2048       & 434M & RoPE  & 2048 & 87.02 & 60.09 & 94.61 & 92.14 & 91.47 & 89.12 & 89.02 & 92.31 & 87.36 \\
\bf GTEv1.5-en-large-8192       & 434M & RoPE  & 8192 & 87.58 & 60.39 & 95.07 & 93.45 & 91.37 & 89.19 & 89.20 & 93.90 & 88.09 \\
\bottomrule
\end{tabular}
}
\caption{
GLUE \cite{wang2018glue} devset scores (w/o WNLI).
$^\alpha$Taken from Table 8 of \citet{liu2019roberta}.
$^\beta$Taken from Table S3 of \citet{portes2023mosaicbert}.
$^\gamma$Taken from Table 2 of \citet{nussbaum2024nomic}.
$^\delta$Taken from Table 2 of \citet{gunther2023jina2}.
The rest of the numbers are from our runs, refer to \S\ref{sec:app:glue} for details.
}
\label{tab:glue-full}
\end{table*}

\section{Text Embedding Evaluation}
We have demonstrated the average scores on MTEB English, Chinese, French and Polish (Table \ref{tab:mteb-multi}).
In this section, we delve into the details, presenting results of different tasks on each language.
For a fair comparison, we do not include the derived models (developed by secondary training on other public off-the-shelf models) in English and Chinese.
In addition to the results obtained from the online leaderboard, our own MTEB evaluations were conducted using version 1.2.0 of \texttt{mteb} library.

\begin{table*}
\centering
\resizebox{\textwidth}{!}{
\begin{tabular}{lcccc|cccccccc}
\toprule[1pt]
\bf MTEB English & \bf Param. & \bf Dim. & \bf Seq. & \bf Avg. &  \bf Class. & \bf Clust. & \bf PairC. & \bf Rerank. & \bf Retr. & \bf STS & \bf Summ.  \\
\bf \#Datasets ($\rightarrow$) &&&& 56 & 12 & 11 & 3 & 4 & 15 & 10 & 1  \\
\midrule
gte-Qwen2-7b-instruct \cite{li2023towards} & 7613M & 3584 & 131072 & 70.24  & 86.58 & 56.92 & 85.79 & 61.42 & 60.25 & 83.04 & 31.35 \\
neural-embedding-v1 & - & - & - & 69.94  & 87.91 & 54.32 & 87.68 & 61.49 & 58.12 & 85.24 & 30.87 \\
NV-Embed-v1  \cite{lee2024nv} & 7851M & 4096 & 32768 & 69.32  & 87.35 & 52.8 & 86.91 & 60.54 & 59.36 & 82.84 & 31.2\\
voyage-large-2-instruct & - & 1024 & 16000 & 68.28  & 81.49 & 53.35 & 89.24 & 60.09 & 58.28 & 84.58 & 30.84\\
gte-Qwen2-1.5B-instruct \cite{li2023towards} & 1776M & 1536 & 131072 & 67.16 & 82.47 & 48.75 & 87.51 & 59.98 & 58.29 & 82.73 & 31.17\\
google-gecko \cite{lee2024gecko} & 1200M & 768 & 2048 & 66.31 & 81.17 & 47.48 & 87.61 & 58.9 & 55.7 & 85.07 & 32.63\\
GritLM-7B \cite{muennighoff2024generative} & 7242M & 4096 & 32768 & 66.76 & 79.46 & 50.61 & 87.16 & 60.49 & 57.41 & 83.35 & 30.37\\
E5-mistral-7b \cite{wang2024improving} & 7111M & 4096 & 32768 & 66.63   & 78.47 & 50.26 & 88.34 & 60.21 & 56.89 & 84.63 & 31.4 \\
text-embedding-3-large & - & 3072 & 8191 & 64.59 & 75.45 & 49.01 & 85.72 & 59.16 & 55.44 & 81.73 & 29.92\\
\midrule
mxbai-embed-large-v1 \cite{emb2024mxbai} & 335M & 1024 & 512 & 64.68 & 75.64 & 46.71 & 87.2 & 60.11 & 54.39 & 85 & 32.71\\
nomic-embed-text-v1 \cite{nussbaum2024nomic} & 137M & 768 & 8192 & 62.39 & 74.12 & 43.91 & 85.15 & 55.69 & 52.81 & 82.06 & 30.08\\
\bf gte-en-large-v1.5 & 434M & 1024 & 8192 & 65.39 & 77.75 & 47.96 & 84.53 & 58.5 & 57.91 & 81.43 & 30.91\\
\bf gte-en-base-v1.5 & 137M & 768 & 8192 & 64.11 & 77.17 & 46.82 & 85.33 & 57.66 & 54.09 & 81.97 & 31.17\\
\midrule
mE5-base \cite{wang2024multilingual} & 278M & 768 & 514 & 59.45 & 73.02 & 37.89 & 83.57 & 54.84 & 48.88 & 80.26 & 30.11\\
mE5-large \cite{wang2024multilingual} & 560M & 1024 & 514 & 61.5 & 74.81 & 41.06 & 84.75 & 55.86 & 51.43 & 81.56 & 29.69\\
BGE-m3 (dense)$^\dagger$ \cite{chen-etal-2024-m3} & 568M & 1024 & 8192 & 59.84 & 74.08 & 37.27 & 84.50 & 55.28 & 48.82 & 81.37 & 31.55 \\
\textbf{\modelname-TRM} (dense) & 305M & 768 & 8192 & 61.40 & 70.89 & 44.31 & 84.23 & 57.47 & 51.08 & 82.11 & 30.58  \\
\midrule
BGE-m3-unsupervised$^\dagger$ \cite{chen-etal-2024-m3} & 560M & 1024 & 8192 & 56.48 & 69.28 & 38.52 & 80.92 & 54.03 & 42.26 & 78.30 & 32.11 \\
\multirow{2}{*}{\bf \modelname-CPT} & \multirow{2}{*}{305M} & \multirow{2}{*}{768} & 512$^*$ & 60.16 & 72.89 & 45.05 & 84.60 & 58.41 & 44.93 & 80.77 & 29.94 \\
 &  &  & 8192 & 60.04 & 72.70 & 45.35 & 84.63 & 58.36 & 44.46 & 80.59 & 30.77 \\
\bottomrule[1pt]
\end{tabular}}
\caption{
Results on MTEB English subset \cite{muennighoff-etal-2023-mteb}. We compare models from the online leaderboard, where derived models (developed by secondary training on other public off-the-shelf models) are not listed.
$^\dagger$Denote our runs.
$^*$To be consistent with the setting in contrastive pre-training, in retrieval tasks, the max sequence length of the document side is set to 1024.
}
\label{tab:mteb-en}
\end{table*}

\begin{table*}
\centering
\resizebox{\textwidth}{!}{
\begin{tabular}{lcccc|ccccccc}
\toprule[1pt]
\bf C-MTEB & \bf Param. & \bf Dim. & \bf Seq. & \bf Avg. &  \bf Class. & \bf Clust. & \bf PairC. & \bf Rerank. & \bf Retr. & \bf STS \\
\bf \#Datasets ($\rightarrow$) &&&& 35 & 9 & 4 & 2 & 4 & 8 & 8  \\
\midrule
gte-Qwen2-7b-instruct \cite{li2023towards} & 7613M & 3584 & 131072 & \bf 72.05 & 75.09 & 66.06 8& 7.48 & 68.92 & 76.03 & 65.33 \\
piccolo-large-zh-v2 \cite{huang2024piccolo2} & - & - & - & 70.95 & 74.59 & 62.17 & 90.24 & 70 & 74.36 & 63.5 \\
OpenSearch-text-hybrid & - & 1792 & 512 & 68.71 & 71.74 & 53.75 & 88.1 & 68.27 & 74.41 & 62.46 \\
Baichuan-text-embedding & - & 1024 & 512 & 68.34 & 72.84 & 56.88 & 82.32 & 69.67 & 73.12 & 60.07 \\
gte-Qwen2-1.5B-instruct \cite{li2023towards} & 1776M & 1536 & 131072 & 67.65 & 71.12 & 54.61 & 86.91 & 68.21 & 71.86 & 60.96 \\
E5-mistral-7b \citep{wang2024improving} & 7111M & 4096 & 32768 & 60.81 & 70.17 & 52.3 & 72.19 & 61.86 & 61.75 & 50.22 \\
\midrule
mE5-base \cite{wang2024multilingual} & 278M & 768 & 514 & 56.21 & 65.35 & 40.68 & 67.07 & 54.35 & 61.63 & 46.49\\
mE5-large \cite{wang2024multilingual} & 560M & 1024 & 514 & 58.81 & 67.34 & 48.23 & 69.89 & 56 & 63.66 & 48.29\\
BGE-m3 (dense)$^\dagger$ \cite{chen-etal-2024-m3} & 568M & 1024 & 8192 & 60.80 & 66.95 & 45.75 & 73.98 & 62.88 & 65.43 & 52.43 \\
\textbf{\modelname-TRM} (dense) & 305M & 768 & 8192 & 62.72 & 64.27 & 47.48 & 78.34 & 68.17 & 71.95 & 52.73 \\
\midrule
BGE-m3-unsupervised$^\dagger$ \cite{chen-etal-2024-m3} & 560M & 1024 & 8192 & 57.53 & 65.04 & 47.10 & 64.09 & 58.14 & 61.45 & 48.42 \\
\multirow{2}{*}{\bf \modelname-CPT} & \multirow{2}{*}{305M} & \multirow{2}{*}{768} & 512$^*$ & 58.67 & 64.64 & 50.21 & 63.95 & 63.77 & 64.23 & 46.74 \\
&&& 8192 & 58.63 & 64.38 & 49.84 & 63.99 & 64.13 & 64.30 & 46.77 \\
\bottomrule[1pt]
\end{tabular}}
\caption{
Results on C-MTEB \cite{xiao2023c} (MTEB Chinese). We compare models from the online leaderboard, where derived models (developed by secondary training on other public off-the-shelf models) are not listed.
$^\dagger$Denote our runs.
$^*$To be consistent with the setting in contrastive pre-training, in retrieval tasks, the max sequence length of the document side is set to 1024.
}
\label{tab:mteb-zh}
\end{table*}

\begin{table*}
\centering
\resizebox{\textwidth}{!}{
\begin{tabular}{lcccc|cccccccc}
\toprule[1pt]
\bf F-MTEB & \bf Param. & \bf Dim. & \bf Seq. & \bf Avg. &  \bf Class. & \bf Clust. & \bf PairC. & \bf Rerank. & \bf Retr. & \bf STS & \bf Summ. \\
\bf \#Datasets ($\rightarrow$) &&&& 26 & 6 & 7 & 2 & 2 & 5 & 3 & 1 \\
\midrule
gte-Qwen2-7b-instruct \cite{li2023towards} & 7613M & 3584 & 131072 & \bf 68.25 & 81.76 & 55.56 & 90.43 & 78.7 & 55.65 & 82.31 & 31.45 \\
gte-Qwen2-1.5B-instruct \cite{li2023towards} & 1776M & 1536 & 131072 & 66.6 & 78.02 & 55.01 & 86.88 & 83.76 & 52.56 & 81.26 & 30.5 \\
voyage-multilingual-2 & - & 1024 & 32000 & 61.65 & 68.56 & 46.57 & 78.66 & 82.59 & 54.56 & 80.13 & 29.96 \\
voyage-law-2 & - & 1024 & 16000 & 60.58 & 68.45 & 44.23 & 77.3 & 82.06 & 52.98 & 80.29 & 30.34 \\
mistral-embed & - & 1024 & - & 59.41 & 68.61 & 44.74 & 77.32 & 80.46 & 46.81 & 79.56 & 31.47 \\
E5-mistral-7b \citep{wang2024improving} & 7111M & 4096 & 32768 & 48.33 & 57.72 & 41.16 & 76.08 & 62.2 & 23.44 & 65.36 & 32.22 \\
\midrule
mE5-base \cite{wang2024multilingual} & 278M & 768 & 514 & 56.19 & 66.8 & 42.66 & 74.82 & 71.76 & 41.19 & 77.22 & 30.76 \\
mE5-large \cite{wang2024multilingual} & 560M & 1024 & 514 & 56.07 & 68.39 & 38.7 & 76.19 & 72.14 & 42.17 & 79.37 & 30.92 \\
BGE-m3 (dense)$^\dagger$ \cite{chen-etal-2024-m3} & 568M & 1024 & 8192 & 58.79 & 71.57 & 36.54 & 79.78 & 77.36 & 51.13 & 80.78 & 31.05  \\
\textbf{\modelname-TRM} (dense) & 305M & 768 & 8192 & 59.79 & 68.72 & 41.66 & 79.47 & 76.47 & 52.97 & 81.36 & 29.74 \\
\midrule
BGE-m3-unsupervised$^\dagger$ \cite{chen-etal-2024-m3} & 560M & 1024 & 8192 & 57.95 & 69.87 & 38.43 & 78.51 & 75.42 & 50.05 & 77.18 & 28.80 \\
\multirow{2}{*}{\bf \modelname-CPT} & \multirow{2}{*}{305M} & \multirow{2}{*}{768} & 512$^*$ & 59.72 & 70.79 & 41.15 & 80.29 & 76.19 & 53.44 & 76.87 & 29.04 \\
&&& 8192 &  59.74 & 70.69 & 41.07 & 79.56 & 77.10 & 53.55 & 77.24 & 28.74 \\
\bottomrule[1pt]
\end{tabular}}
\caption{
Results on F-MTEB \cite{ciancone2024mteb-french} (MTEB French). We compare top-performing models from the online leaderboard.
$^\dagger$Denote our runs.
$^*$To be consistent with the setting in contrastive pre-training, in retrieval tasks, the max sequence length of the document side is set to 1024.
}
\label{tab:mteb-fr}
\end{table*}

\begin{table*}
\centering
\resizebox{\textwidth}{!}{
\begin{tabular}{lcccc|cccccc}
\toprule[1pt]
\bf MTEB Polish & \bf Param. & \bf Dim. & \bf Seq. & \bf Avg. &  \bf Class. & \bf Clust. & \bf PairClass. & \bf Retr. & \bf STS \\
\bf \#Datasets ($\rightarrow$) &&&& 26 & 7 & 1 & 4 & 11 & 3 \\
\midrule
gte-Qwen2-7b-instruct \cite{li2023towards} & 7613M & 3584 & 131072 & \bf 67.86 & 77.84 & 51.36 & 88.48 & 54.69 & 70.86 \\
gte-Qwen2-1.5B-instruct \cite{li2023towards} & 1776M & 1536 & 131072 & 64.04 & 72.29 & 44.59 & 84.87 & 51.88 & 68.12\\
mmlw-roberta-large \cite{dadas2024pirb} & 435M & 1024 & 514 & 63.23 & 66.39 & 31.16 & 89.13 & 52.71 & 70.59 \\
mmlw-e5-large  \cite{dadas2024pirb} & 560M & 1024 & 514 & 61.17 & 61.07 & 30.62 & 85.9 & 52.63 & 69.98\\
mmlw-roberta-base  \cite{dadas2024pirb} & 124M & 768 & 514 & 61.05 & 62.92 & 33.08 & 88.14 & 49.92 & 70.7\\
\midrule
mE5-base \cite{wang2024multilingual} & 278M & 768 & 514 & 55.62 & 59.01 & 24.97 & 82.15 & 44.01 & 65.13\\
mE5-large \cite{wang2024multilingual} & 560M & 1024 & 514 & 60.08 & 63.82 & 33.88 & 85.5 & 48.98 & 66.91\\
BGE-m3 (dense)$^\dagger$ \cite{chen-etal-2024-m3} & 568M & 1024 & 8192 & 60.35 & 65.15 & 25.21 & 86.46 & 48.51 & 69.44 \\
\textbf{\modelname-TRM} (dense) & 305M & 768 & 8192 & 58.22 & 60.15 & 33.67 & 85.45 & 46.40 & 68.92 \\
\midrule
BGE-m3-unsupervised$^\dagger$ \cite{chen-etal-2024-m3} & 560M & 1024 & 8192 & 55.98 & 60.30 & 40.17 & 79.01 & 43.26 & 67.05 \\
\multirow{2}{*}{\bf \modelname-CPT} & \multirow{2}{*}{305M} & \multirow{2}{*}{768} & 512$^*$ & 57.66 & 62.72 & 38.04 & 79.70 & 45.55 & 67.39 \\
&&& 8192 &  57.11 & 61.55 & 38.15 & 79.53 & 45.29 & 66.53 \\
\bottomrule[1pt]
\end{tabular}}
\caption{
Results on MTEB Polish subset \cite{poswiata2024pl} We compare top-performing models from the online leaderboard.
$^\dagger$Denote our runs.
$^*$To be consistent with the setting in contrastive pre-training, in retrieval tasks, the max sequence length of the document side is set to 1024.
}
\label{tab:mteb-pl}
\end{table*}

\paragraph{MTEB-en}
Table \ref{tab:mteb-en} shows the results on English MTEB \cite{muennighoff-etal-2023-mteb}.
For reference, we include our English embedding models (\texttt{Our-en-base/large-embed}, trained by the two-stage contrastive learning on the English part of our data) and top-performing systems from the online leaderboard.
We can see that the multilingual models still have a noticeable gap compared to the English models.

\paragraph{MTEB-zh}
Table \ref{tab:mteb-zh} presents the C-MTEB \cite{xiao2023c} (MTEB Chinese subset) results.
We include the results of several LLM-based embedding models and APIs.
Given that the Chinese community is also keen on optimizing embedding models, the gap between multilingual models and Chinese models is quite noticeable.

\paragraph{MTEB-fr}
Table \ref{tab:mteb-fr} demonstrates the F-MTEB \cite{ciancone2024mteb-french} (MTEB French subset) results.
Our TRM dense is comparable to the specialized French API \texttt{mistral-embed}.
However, compared to our our-cpt model, the improvement from fine-tuning is not significant.

\paragraph{MTEB-pl}
Table \ref{tab:mteb-pl} lists the Polish MTEB \cite{poswiata2024pl} results.
Our model does not outperform large-sized BGE and mE5.
We speculate this may be due to the limited amount of Polish pairs in the contrastive pre-training, resulting in insufficient training.

\section{Text Retrieval Evaluation}\label{sec:app:retrieval}
The retrieval process can be divided into two main stages: recall and reranking. In the recall stage, documents are retrieved using both dense vectors and sparse representations. The final recall score is calculated by weighting the dense retrieval score with a fixed coefficient of 1 and the sparse retrieval score with coefficients ranging from 0.001 to 0.01. Documents not retrieved by either method receive a score of 0. During the ranking stage, the top 100 documents from the recall results are selected as candidates. These candidates are then sorted using our reranker model to produce the final retrieval results.

We present the detail results of MLDR \cite{chen-etal-2024-m3} (multilingual long-context retrieval, Table \ref{tab:bge_long_results}), MKQA \cite{longpre2021mkqa} (multilingual, Table \ref{tab:mkqa_recall@20_results}), MIRACL \cite{10.1162/tacl_a_00595} (multilingual, Table \ref{tab:miracl_ndcg_results}, BEIR \cite{thakur2beir} (English, Table \ref{tab:beir_results}) and LoCo \cite{saadbenchmarking} (English long-context, Table \ref{tab:loco}).

\begin{table*}
\centering
\setlength{\tabcolsep}{2.5pt}
\resizebox{\textwidth}{!}{
\begin{tabular}{lc|c|ccccccccccccc}
\toprule
& Max Length & Avg & ar & de & en & es & fr & hi & it & ja & ko & pt & ru & th & zh \\
\midrule
BM25 & 8192 & 53.6 & 45.1 & 52.6 & 57.0 & 78.0 & 75.7 & 43.7 & 70.9 & 36.2 & 25.7 & 82.6 & 61.3 & 33.6 & 34.6 \\
mE5$_{\mathrm{{\text{large}}}}$ & 512 & 34.2 & 33.0 & 26.9 & 33.0 & 51.1 & 49.5 & 21.0 & 43.1 & 29.9 & 27.1 & 58.7 & 42.4 & 15.9 & 13.2 \\
mE5$_{\mathrm{{\text{base}}}}$ & 512 &  30.5 & 29.6 & 26.3 & 29.2 & 45.2 & 46.7 & 19.0 & 40.9 & 24.9 & 20.9 & 50.8 & 37.8 & 12.2 & 12.8 \\
E5$_{\mathrm{\text{mistral-7b}}}$ & 8192 & 42.6 & 29.6 & 40.6 & 43.3 & 70.2 & 60.5 & 23.2 & 55.3 & 41.6 & 32.7 & 69.5 & 52.4 & 18.2 & 16.8 \\
BGE-m3-Dense           & 8192 & 52.5 & 47.6 & 46.1 & 48.9 & 74.8 & 73.8 & 40.7 & 62.7 & 50.9 & 42.9 & 74.4 & 59.5 & 33.6 & 26.0 \\
BGE-m3-Sparse          & 8192 & 62.2 & 58.7 & 53.0 & 62.1 & 87.4 & 82.7 & 49.6 & 74.7 & 53.9 & 47.9 & 85.2 & 72.9 & 40.3 & 40.5 \\
BGE-m3-Dense+Sparse    & 8192 & 64.8 & 63.0 & 56.4 & 64.2 & 88.7 & 84.2 & 52.3 & 75.8 & 58.5 & 53.1 & 86.0 & 75.6 & 42.9 & 42.0 \\
\midrule
\textbf{\modelname-TRM} Dense & 8192 & 56.6 & 55.0 & 54.9 & 51.0 & 81.2 & 76.2 & 45.2 & 66.7 & 52.1 & 46.7 & 79.1 & 64.2 & 35.3 & 27.4  \\
\textbf{\modelname-TRM} Sparse & 8192 & 71.0 & 74.3 & 66.2 & 66.4 & 93.6 & 88.4 & 61.0 & 82.2 & 66.2 & 64.2 & 89.9 & 82.0 & 47.4 & 41.8 \\
\textbf{\modelname-TRM} Dense+Sparse & 8192 & 71.3 & 74.6 & 66.6 & 66.5 & 93.6 & 88.6 & 61.6 & 83.0 & 66.7 & 64.6 & 89.8 & 82.1 & 47.7 & 41.4 \\
+ \textbf{\modelname-reranker} & 8192 & 73.8 & 76.6 & 70.4 & 69.3 & 96.4 & 89.6 & 67.8 & 81.9 & 68.1 & 71.1 & 90.2 & 86.1 & 46.7 & 44.8 \\
\bottomrule
\end{tabular}}
\caption{Evaluation of multilingual long-doc retrieval on the MLDR \cite{chen-etal-2024-m3} testset (measured by nDCG@10).}
\label{tab:bge_long_results}
\end{table*}

\begin{table*}[!t]
\centering
\setlength{\tabcolsep}{3pt}
\resizebox{\textwidth}{!}{
\begin{tabular}{lccccccc|ccccc|ccc|c}
\hline
& \multicolumn{7}{c|}{Baselines} & \multicolumn{5}{c|}{M3-Embedding} & \multicolumn{3}{c|}{\bf \modelname-TRM} & \bf \modelname-reranker\\
\hline
& BM25 & mDPR & mContriever & mE5$_{\mathrm{{\text{large}}}}$ & mE5$_{\mathrm{\text{base}}}$ & E5$_{\mathrm{\text{mistral-7b}}}$ & OpenAI-3 & Dense & Sparse & Multi-vec & D+S & All & Dense & Sparse & D+S & ReRank\\
\hline
ar      & 13.4 & 33.8 & 43.8 & 59.7 & 44.3 & 47.6 & 55.1 & 61.9 & 19.5 & 62.6 & 61.9 & \textbf{63.0} & 55.9 & 17.5 & 56.0 & 58.2\\
da      & 36.2 & 55.7 & 63.3 & 71.7 & 63.6 & \textbf{72.3} & 67.6 & 71.2 & 45.1 & 71.7 & 71.3 & 72.0 & 69.8 & 37.9 & 69.7 & 71.0\\
de      & 23.3 & 53.2 & 60.2 & \textbf{71.2} & 62.3 & 70.8 & 67.6 & 69.8 & 33.2 & 69.6 & 70.2 & 70.4 & 68.9 & 27.0 & 69.1 & 70.1\\
es      & 29.8 & 55.4 & 62.3 & 70.8 & 63.8  & \textbf{71.6} & 68.0 & 69.8 & 40.3 & 70.3 & 70.2 & 70.7 & 69.6 & 35.1 & 70.0 & 71.0\\
fi      & 33.2 & 42.8 & 58.7 & 67.7 & 53.0 & 63.6 & 65.5 & 67.8 & 41.2 & 68.3 & 68.4 & \textbf{68.9} & 64.2 & 35.3 & 64.5 & 64.9\\
fr      & 30.3 & 56.5 & 62.6 & 69.5 & 61.2 & \textbf{72.7} & 68.2 & 69.6 & 43.2 & 70.1 & 70.1 & 70.8 & 69.8 & 36.9 & 70.4 & 71.0 \\
he      & 16.1 & 34.0 & 50.5 & 61.4 & 37.4 & 32.4 & 46.3 & 63.4 & 24.5 & 64.4 & 63.5 & \textbf{64.6} & 55.4 & 22.0 & 55.4 & 56.5 \\
hu      & 26.1 & 46.1 & 57.1 & 68.0 & 55.9 & \textbf{68.3} & 64.0 & 67.1 & 34.5 & 67.3 & 67.7 & 67.9 & 64.6 & 28.8 & 65.0 & 66.1 \\
it      & 31.5 & 53.8 & 62.0 & 71.2 & 61.6 & \textbf{71.3} & 67.6 & 69.7 & 41.5 & 69.9 & 69.9 & 70.3 & 69.0 & 36.2 & 69.2 & 70.1 \\
ja      & 14.5 & 46.3 & 50.7 & 63.1 & 51.7 & 57.6 & 64.2 & 67.0 & 23.3 & 67.8 & 67.1 & \textbf{67.9} & 65.3 & 19.5 & 65.2 & 67.2 \\
km      & 20.7 & 20.6 & 18.7 & 18.3 & 28.2 & 23.3 & 25.7 & 58.5 & 24.4 & 59.2 & 58.9 & \textbf{59.5} & 53.6 & 21.9 & 53.8 & 54.7 \\
ko      & 18.3 & 36.8 & 44.9 & 58.9 & 40.4 & 49.4 & 53.9 & 61.9 & 24.3 & 63.2 & 62.1 & \textbf{63.3} & 55.9 & 21.4 & 56.1 & 58.9 \\
ms      & 42.3 & 53.8 & 63.7 & 70.2 & 62.4 & 71.1 & 66.1 & 71.6 & 52.5 & 72.1 & 71.8 & \textbf{72.3} & 69.9 & 47.8 & 70.2 & 70.9 \\
nl      & 42.5 & 56.9 & 63.9 & \textbf{73.0} & 65.0 & 74.5 & 68.8 & 71.3 & 52.9 & 71.8 & 71.7 & 72.3 & 70.7 & 47.4 & 70.9 & 71.5 \\
no      & 38.5 & 55.2 & 63.0 & 71.1 & 62.0 & 70.8 & 67.0 & 70.7 & 47.0 & 71.4 & 71.1 & \textbf{71.6} & 69.1 & 39.7 & 69.2 & 70.2 \\
pl      & 28.7 & 50.4 & 60.9 & 70.5 & 57.2 & \textbf{71.5} & 66.1 & 69.4 & 36.4 & 70.0 & 69.9 & 70.4 & 68.4 & 31.4 & 68.3 & 69.6 \\
pt      & 31.8 & 52.5 & 61.0 & 66.8 & 58.7 & \textbf{71.6} & 67.7 & 69.3 & 40.2 & 70.0 & 69.8 & 70.6 & 69.6 & 34.9 & 69.6 & 70.7 \\
ru      & 21.8 & 49.8 & 57.9 & \textbf{70.6} & 58.7 & 68.7 & 65.1 & 69.4 & 29.2 & 70.0 & 69.4 & 70.0 & 68.5 & 25.8 & 68.5 & 69.6 \\
sv      & 41.1 & 54.9 & 62.7 & 72.0 & 61.3 & \textbf{73.3} & 67.8 & 70.5 & 49.8 & 71.3 & 71.5 & 71.5 & 69.5 & 43.3 & 69.9 & 70.6 \\
th      & 28.4 & 40.9 & 54.4 & 69.7 & 59.7 & 57.1 & 55.2 & 69.6 & 34.7 & 70.5 & 69.8 & \textbf{70.8} & 65.0 & 30.6 & 65.2 & 66.9\\
tr      & 33.5 & 45.5 & 59.9 & 67.3 & 59.2 & 65.5 & 64.9 & 68.2 & 40.9 & 69.0 & 69.1 & \textbf{69.6} & 67.7 & 36.0 & 67.7 & 69.0 \\
vi      & 33.6 & 51.3 & 59.9 & 68.7 & 60.0 & 62.3 & 63.5 & 69.6 & 42.2 & 70.5 & 70.2 & \textbf{70.9} & 69.4 & 37.6 & 69.3 & 70.3\\
zh\_cn  & 19.4 & 50.1 & 55.9 & 44.3 & 38.3 & 61.2 & 62.7 & 66.4 & 26.9 & 66.7 & 66.6 & \textbf{67.3} & 68.2 & 23.2 & 68.4 & 69.5\\
zh\_hk  & 23.9 & 50.2 & 55.5 & 46.4 & 38.3 & 55.9 & 61.4 & 65.8 & 31.2 & 66.4 & 65.9 & \textbf{66.7} & 63.7 & 27.8 & 63.8 & 65.8 \\
zh\_tw  & 22.5 & 50.6 & 55.2 & 45.9 & 39.0 & 56.5 & 61.6 & 64.8 & 29.8 & 65.3 & 64.9 & \textbf{65.6} & 63.8 & 26.6 & 63.9 & 65.7 \\
\hline
Avg     & 28.1 & 47.9 & 56.3 & 63.5 & 53.7 & 62.4 & 62.1 & 67.8 & 36.3 & 68.4 & 68.1 & \textbf{68.8} & 65.8 & 31.6 & 66.0 & 67.2\\
\hline
\end{tabular}
}
\caption{
Recall@20 on MKQA \cite{longpre2021mkqa} dataset for cross-lingual retrieval in all 25 languages.
The \texttt{All} of M3-Embedding denotes the hybrid retrieval result of dense, sparse, and multi-vec scores.
}
\label{tab:mkqa_recall@20_results}
\end{table*}

\begin{table*}[!t]
\centering
\footnotesize
\setlength{\tabcolsep}{2pt}
\resizebox{\textwidth}{!}{
\begin{tabular}{l|c|cccccccccccccccccc}
\toprule
Model & Avg & ar & bn & en & es & fa & fi & fr & hi & id & ja & ko & ru & sw & te & th & zh & de & yo \\
\midrule
BM25 & 31.9 & 39.5 & 48.2 & 26.7 & 7.7 & 28.7 & 45.8 & 11.5 & 35.0 & 29.7 & 31.2 & 37.1 & 25.6 & 35.1 & 38.3 & 49.1 & 17.5 & 12.0 & 56.1 \\
mE5$_{\mathrm{{\text{large}}}}$ & 65.4 & 76.0 & 75.9 & 52.9 & 52.9 & 59.0 & 77.8 & 54.5 & 62.0 & 52.9 & 70.6 & 66.5 & 67.4 & 74.9 & 84.6 & 80.2 & 56.0 & 56.4 & 56.5 \\
mE5$_{\mathrm{\text{base}}}$ & 60.13 & 71.6 & 70.2 & 51.2 & 51.5 & 57.4 & 74.4 & 49.7 & 58.4 & 51.1 & 64.7 & 62.2 & 61.5 & 71.1 & 75.2 & 75.2 & 51.5 & 43.4 & 42.3 \\
E5$_{\mathrm{\text{mistral-7b}}}$ & 62.2 & 73.3 & 70.3 & 57.3 & 52.2 & 52.1 & 74.7 & 55.2 & 52.1 & 52.7 & 66.8 & 61.8 & 67.7 & 68.4 & 73.9 & 74.0 & 54.0 & 54.0 & 58.8 \\
OpenAI-3 & 54.9 & - & - & - & - & - & - & - & - & - & - & - & - & - & - & - & - & - & - \\
BGE-M3-Dense & 67.8 & 78.4 & 80.0 & 56.9 & 55.5 & 57.7 & 78.6 & 57.8 & 59.3 & 56.0 & 72.8 & 69.9 & 70.1 & 78.6 & 86.2 & 82.6 & 61.7 & 56.8 & 60.7 \\
BGE-M3-Sparse & 53.9 & 67.1 & 68.7 & 43.7 & 38.8 & 45.2 & 65.3 & 35.5 & 48.2 & 48.9 & 56.3 & 61.5 & 44.5 & 57.9 & 79.0 & 70.9 & 36.3 & 32.2 & 70.0 \\
BGE-M3-Multi-vec & 69.0 & 79.6 & 81.1 & 59.4 & 57.2 & 58.8 & 80.1 & 59.0 & 61.4 & 58.2 & 74.5 & 71.2 & 71.2 & 79.0 & 87.9 & 83.0 & 62.7 & 57.9 & 60.4 \\
BGE-M3-Dense+Sparse & 68.9 & 79.6 & 80.7 & 58.8 & 57.5 & 59.2 & 79.7 & 57.6 & 62.8 & 58.3 & 73.9 & 71.3 & 69.8 & 78.5 & 87.2 & 83.1 & 62.5 & 57.6 & \textbf{61.8} \\
BGE-M3 All & \textbf{70.0} & \textbf{80.2} & \textbf{81.5} & \textbf{59.8} & \textbf{59.2} & \textbf{60.3} & \textbf{80.4} & \textbf{60.7} & \textbf{63.2} & \textbf{59.1} & \textbf{75.2} & \textbf{72.2} & \textbf{71.7} & \textbf{79.6} & \textbf{88.2} & \textbf{83.8} & \textbf{63.9} & \textbf{59.8} & 61.5 \\
\midrule
\textbf{\modelname-TRM} Dense & 62.1 & 71.4 & 72.7 & 54.1 & 51.4 & 51.2 & 73.5 & 53.9 & 51.6 & 50.3 & 65.8 & 62.7 & 63.2 & 69.9 & 83.0 & 74.0 & 60.8 & 49.7 & 58.3 \\
\textbf{\modelname-TRM} Sparse & 55.9 & 66.5 & 70.4 & 35.6 & 46.2 & 40.0 & 47.6 & 66.5 & 39.8 & 48.9 & 47.9 & 59.3 & 64.3 & 47.1 & 59.4 & 83.0 & 70.5 & 73.7 & 39.9\\
\textbf{\modelname-TRM} Dense+Sparse & 63.5 & 73.4 & 75.1 & 49.9 & 57.6 & 62.7 & 52.0 & 74.7 & 53.5 & 56.4 & 52.8 & 67.1 & 66.7 & 63.5 & 69.5 & 85.2 & 75.8 & 58.4 & 58.8\\
+ \textbf{\modelname-reranker} & 68.5 & 77.1 & 63.1 & 78.6 & 56.3 & 72.4 & 80.3 & 79.6 & 58.6 & 59.1 & 74.6 & 75.5 & 59.4 & 56.3 & 56.5 & 62.2 & 72.2 & 86.3 & 65.1 \\
\hline
\end{tabular}}
\caption{Multi-lingual retrieval performance on the MIRACL \cite{10.1162/tacl_a_00595} dev set (measured by nDCG@10).}
\label{tab:miracl_ndcg_results}
\end{table*}

\begin{table*}
\centering
\setlength{\tabcolsep}{2pt}
\resizebox{\textwidth}{!}{
\begin{tabular}{lc|cccccccccccccccc}
\toprule
BEIR &
Avg. &
\begin{tabular}{c} Argu- \\ Ana \end{tabular} &
\begin{tabular}{c} Cli-\\mate- \\ Fever \end{tabular} &
\begin{tabular}{c} CQA-\\Dup- \\ Stack \end{tabular} &
\begin{tabular}{c} DB- \\ Pedia \end{tabular} &
Fever&
FiQA &
\begin{tabular}{c} Hotpot- \\ QA \end{tabular} &
\begin{tabular}{c} MS \\ MAR-\\CO \end{tabular} & 
\begin{tabular}{c} NF- \\ Corpus \end{tabular} & 
NQ &
Quora &
\begin{tabular}{c} Sci- \\ docs \end{tabular} &
\begin{tabular}{c} Sci- \\ fact \end{tabular} &
\begin{tabular}{c} Touche- \\ 2020 \end{tabular} &
\begin{tabular}{c} Trec- \\ Covid \end{tabular} &
\\ \midrule
gte-Qwen2-7B-instruct & 60.25 & 64.27 & 45.88 & 46.43 & 52.42 & 95.11 & 62.03 & 73.08 & 45.98 & 40.6 & 67 & 90.09 & 28.91 & 79.06 & 30.57 & 82.26 \\
NV-Embed-v1 & 59.36 & 68.2 & 34.72 & 50.51 & 48.29 & 87.77 & 63.1 & 79.92 & 46.49 & 38.04 & 71.22 & 89.21 & 20.19 & 78.43 & 28.38 & 85.88 \\
gte-Qwen2-1.5B-instruct & 58.29 & 69.72 & 42.91 & 44.76 & 48.69 & 91.57 & 54.7 & 68.95 & 43.36 & 39.34 & 64 & 89.64 & 24.98 & 78.44 & 27.89 & 85.38 \\
voyage-large-2-instruct & 58.28 & 64.06 & 32.65 & 46.6 & 46.03 & 91.47 & 59.76 & 70.86 & 40.6 & 40.32 & 65.92 & 87.4 & 24.32 & 79.99 & 39.16 & 85.07 \\
neural-embedding-v1 & 58.12 & 67.21 & 32.3 & 49.11 & 48.05 & 89.46 & 58.94 & 78.87 & 42 & 42.6 & 68.36 & 89.02 & 27.69 & 78.82 & 24.06 & 75.33 \\
GritLM-7B & 57.41 & 63.24 & 30.91 & 49.42 & 46.6 & 82.74 & 59.95 & 79.4 & 41.96 & 40.89 & 70.3 & 89.47 & 24.41 & 79.17 & 27.93 & 74.8 \\
e5-mistral-7b-instruct & 56.89 & 61.88 & 38.35 & 42.97 & 48.89 & 87.84 & 56.59 & 75.72 & 43.06 & 38.62 & 63.53 & 89.61 & 16.3 & 76.41 & 26.39 & 87.25 \\
google-gecko & 55.7 & 62.18 & 33.21 & 48.89 & 47.12 & 86.96 & 59.24 & 71.33 & 32.58 & 40.33 & 61.28 & 88.18 & 20.34 & 75.42 & 25.86 & 82.62 \\
text-embedding-3-large & 55.44 & 58.05 & 30.27 & 47.54 & 44.76 & 87.94 & 55 & 71.58 & 40.24 & 42.07 & 61.27 & 89.05 & 23.11 & 77.77 & 23.35 & 79.56 \\
\midrule
\bf gte-en-large-v1.5 & 57.91 & 72.11 & 48.36 & 42.16 & 46.3 & 93.81 & 63.23 & 68.18 & 42.93 & 36.95 & 56.08 & 89.67 & 26.35 & 82.43 & 22.55 & 77.49 \\
\bf gte-en-base-v1.5 & 54.09 & 63.49 & 40.36 & 39.52 & 39.9 & 94.81 & 48.65 & 67.75 & 42.62 & 35.88 & 52.96 & 88.42 & 21.92 & 76.77 & 25.22 & 73.13 \\
\midrule \midrule
BM25 & 41.7 & 31.5 & 21.3 & 29.9 & 31.3 & 75.3 & 23.6 & 60.3 & 22.8 & 32.5 & 32.9 & 78.9 & 15.8 & 66.5 & 36.7 & 65.6 \\
mE5-large & 51.43 & 54.38 & 25.73 & 39.68 & 41.29 & 82.81 & 43.8 & 71.23 & 43.7 & 33.99 & 64.06 & 88.18 & 17.47 & 70.41 & 23.39 & 71.33 \\
mE5-base & 48.88 & 44.23 & 23.86 & 38.52 & 40.36 & 79.44 & 38.17 & 68.56 & 42.27 & 32.46 & 60.02 & 87.65 & 17.16 & 69.35 & 21.35 & 69.76 \\
\midrule
BGE-M3 Dense$^\dagger$ & 48.34 & 53.95 & 29.52 & 39.09 & 39.80 & 81.38 & 41.30 & 69.44 & 38.32 & 31.43 & 60.60 & 88.57 & 16.39 & 64.36 & 22.63 & 55.59 \\
BGE-M3 Sparse$^\dagger$ & 38.30 & 25.08 & 24.69 & 27.51 & 23.21 & 88.36 & 26.79 & 68.45 & 19.59 & 27.5 & 17.98 & 73.82 & 8.89 & 64.37 & 30.26 & 48.00 \\
BGE-M3 Dense+Sparse$^\dagger$ & 49.41 & 53.88 & 30.21 & 39.10 & 39.89 & 81.24 & 40.25 & 70.11 & 37.62 & 32.53 & 59.58 & 88.62 & 15.59 & 65.74 & 31.12 & 55.67 \\
\midrule
\textbf{\modelname-TRM} Dense & 51.07 & 58.36 & 34.83 & 38.12 & 40.11 & 92.07 & 44.99 & 63.03 & 39.92 & 36.66 & 58.10 & 88.02 & 18.26 & 73.42 & 22.76 & 57.4 \\
\textbf{\modelname-TRM} Sparse & 39.24 & 40.06 & 24.17 & 25.11 & 20.0 & 88.32 & 28.58 & 64.68 & 19.39 & 28.34 & 19.71 & 76.84 & 10.92 & 67.72 & 21.52 & 53.33 \\
\textbf{\modelname-TRM} Dense+Sparse & 51.43 & 58.48 & 34.89 & 38.36 & 39.72 & 93.14 & 44.98 & 65.01 & 39.99 & 36.67 & 56.90 & 89.05 & 18.26 & 73.45 & 24.09 & 58.46\\
+ \textbf{\modelname-reranker} & 55.42 & 58.53 & 44.93 & 38.37 & 45.62 & 93.9 & 44.38 & 74.51 & 44.99 & 36.29 & 65.21 & 81.67 & 18.42 & 75.59 & 31.29 & 77.75 \\
\midrule
BGE-M3-unsupervised$^\dagger$ & 42.26 & 59.07 & 23.05 & 38.10 & 31.16 & 59.15 & 36.57 & 53.39 & 27.79 & 30.67 & 39.69 & 86.38 & 15.08 & 61.26 & 17.62 & 54.90 \\
\textbf{\modelname-CPT}-512,1024 & 44.93 & 52.99 & 17.93 & 45.01 & 37.63 & 34.13 & 48.38 & 54.39 & 31.76 & 39.01 & 48.48 & 86.82 & 22.95 & 72.46 & 18.56 & 63.46 \\
\textbf{\modelname-CPT}-8192 & 44.46 & 55.14 & 15.85 & 44.73 & 38.74 & 27.42 & 47.45 & 55.93 & 31.79 & 38.62 & 49.27 & 86.81 & 22.72 & 73.08 & 17.08 & 62.27 \\
\bottomrule
\end{tabular}
}
\caption{
BEIR benchmark~\citep{thakur2beir} nDCG@10 scores. We include top models from MTEB Retrieval English leaderboard.
$^\dagger$Denote our runs.
}
\label{tab:beir_results}
\end{table*}

\begin{table*}[htb]
\centering
\resizebox{\textwidth}{!}{
\begin{tabular}{lcccc|cccccccc}
\toprule
\bf Model & \bf Param. & \bf Dim. & \bf Seq & \bf Avg. & Tau Scr. & Tau Gov. & Tau QMS. & QASP. Tit. Art. & QASP. Abs. Art. \\
\midrule
Jina$_\texttt{base-v2}$~\citep{gunther2023jina2} & 137M & 768 & 8192 & 85.5 & 93.3 & 98.6 & 40.8 & 95.1 & 99.3 \\
nomic-embed-text-v1 \cite{nussbaum2024nomic} & 137M & 768 & 8192 & 85.5 & 90.9 & 97.8 & 44.2 & 94.9 & 99.9 \\
text-embedding-3-small & - & 1536 & 8192 & 82.4 & 92.2 & 97.7 & 27.4 & 95.9 & 98.9 \\
text-embedding-3-large & - & 3072 & 8192 & 79.4 & 88.0 & 93.6 & 25.5 & 93.2 & 96.8 \\
\bf \modelname-en-base-embed  & 137M & 768  & 8192 & 87.4 & 91.8 & 98.6 & 49.9 & 97.1 & 99.8 \\
\bf \modelname-en-large-embed & 434M & 1024 & 8192 & 86.7 & 92.6 & 98.7 & 44.5 & 97.8 & 99.8 \\
mE5$_{\mathrm{{\text{base}}}}$ \cite{wang2024multilingual} & 279M & 768 & 512 & 72.2 & 68.9 & 87.6 & 30.5 & 85.1 & 88.9 \\
mE5$_{\mathrm{{\text{large}}}}$ \cite{wang2024multilingual} & 279M & 1024 & 512 & 74.3 & 70.4 & 89.5 & 37.6 & 89.5 & 85.4\\
E5$_\texttt{mistral}$~\citep{wang2024improving} & 7B & 4096 & 4096  & 87.8 & 95.9 & 98.3 & 46.8 & 98.4 & 99.8 \\
BGE-M3-Dense$^\dagger$ \cite{chen-etal-2024-m3} & 568M & 1024 & 8192 &  84.9 & 93.8 & 97.4 & 41.9 & 93.2 & 98.3 \\
BGE-M3-Sparse$^\dagger$ \cite{chen-etal-2024-m3} & 568M & 1024 & 8192 & 84.9  & 95.5 & 97.9 & 46.7 & 85.7 & 98.9 \\
BGE-M3-Dense+Sparse$^\dagger$ \cite{chen-etal-2024-m3} & 568M & 1024 & 8192 & 87.4 & 97.7 & 98.2 & 47.7 & 93.6 & 99.7 \\
\midrule
\textbf{\modelname-TRM} Dense & 434M & 768 & 8192 & 88.9  & 95.1 & 97.7 & 58.5 & 94.6 & 98.7 \\
\textbf{\modelname-TRM} Sparse & 434M & 768 & 8192 & 88.1 & 97.6 & 97.9 & 60.1 & 85.5 & 99.2 \\
\textbf{\modelname-TRM} Dense+Sparse & 434M & 768 & 8192 & 91.3 & 98.2 & 98.3 & 66.5 & 94.6 & 98.7 \\
\bottomrule
\end{tabular}
}
\caption{
The nCDG@10 scores on the LoCo benchmark \cite{saadbenchmarking}.
$^\dagger$Denote our runs.
}
\label{tab:loco}
\end{table*}

\end{CJK}
\end{document}